\documentclass[preprint,12pt,authoryear]{elsarticle}
\usepackage{amssymb}
\usepackage{amsmath}
\usepackage{orcidlink}
\usepackage{hyperref}
\usepackage{float}
\usepackage[skip=8pt, justification=centering]{caption}
\usepackage{enumitem}
\usepackage{multirow}
\usepackage{array}
\usepackage{booktabs}
\usepackage{subcaption}
\usepackage{tabularx}
\usepackage{longtable}
\usepackage{rotating}
\usepackage{adjustbox}
\usepackage{xcolor}
\usepackage[utf8]{inputenc}
\usepackage[T5]{fontenc}
\usepackage{caption}
\usepackage{color}
\usepackage{pdflscape}

\begin{document}

\begin{frontmatter}

\title{Multi-objective hybrid knowledge distillation for efficient deep learning in smart agriculture}

\author{Phi-Hung Hoang}
\ead{hunghpde180523@fpt.edu.vn}

\author{Nam-Thuan Trinh}
\ead{thuantnde180305@fpt.edu.vn}

\author{Van-Manh Tran}
\ead{manhtvde180090@fpt.edu.vn}

\author{Thi-Thu-Hong Phan\corref{cor}\orcidlink{0000-0001-6880-3721}}
\ead{hongptt11@fe.edu.vn}

\cortext[cor]{Corresponding author}

\hypersetup{pdfauthor={Thi-Thu-Hong Phan}}

\affiliation{organization={Department of Artificial Intelligence, FPT University},
            addressline={Da Nang},
            postcode={550000}, 
            country={Vietnam}}

\begin{abstract}

Deploying deep learning models on resource-constrained edge devices remains a major challenge in smart agriculture due to the trade-off between computational efficiency and recognition accuracy. To address this challenge, this study proposes a hybrid knowledge distillation framework for developing a lightweight yet high-performance convolutional neural network. The proposed approach designs a customized student model that combines inverted residual blocks with dense connectivity and trains it under the guidance of a ResNet18 teacher network using a multi-objective strategy that integrates hard-label supervision, feature-level distillation, response-level distillation, and self-distillation.
Experiments are conducted on a rice seed variety identification dataset containing nine varieties and further extended to four plant leaf disease datasets, including rice, potato, coffee, and corn, to evaluate generalization capability. On the rice seed variety classification task, the distilled student model achieves an accuracy of 98.56\%, which is only 0.09\% lower than the teacher model (98.65\%), while requiring only 0.68 GFLOPs and approximately 1.07 million parameters. This corresponds to a reduction of about 2.7 times in computational cost and more than 10 times in model size compared with the ResNet18 teacher model. In addition, compared with representative pretrained models, the proposed student reduces the number of parameters by more than 6 times relative to DenseNet121 and by over 80 times compared with the Vision Transformer (ViT) architecture, while maintaining comparable or superior classification accuracy. Consistent performance gains across multiple plant leaf disease datasets further demonstrate the robustness, efficiency, and strong deployment potential of the proposed framework for hardware-limited smart agriculture systems.

\end{abstract}


\end{frontmatter}

\section{Introduction}

Smart agriculture has emerged as a key contributor to global food security, enabling improvements in crop productivity while reducing biological risks~\citep{Kumar2024, Raj2025, Roy2025}. With the development of precision agriculture, real-time monitoring tasks such as rice variety identification and plant leaf disease classification have become increasingly important. Traditionally, these tasks rely heavily on the visual experience of agronomists, which is time consuming, highly subjective, and difficult to scale under large scale farming conditions \citep{Shafik2025PlantDiseaseDL}. Therefore, the development of automated computer vision systems capable of fast and accurate diagnosis is a necessary step toward improving the efficiency of agricultural management.

In the initial stage of automated agricultural image analysis, researchers predominantly adopted traditional machine learning methods with handcrafted features, including color descriptors, Local Binary Patterns (LBP), and Gray-Level Co-occurrence Matrix (GLCM)~\citep{Ahmed2021, Rusli2022, Alsakar2024, Le2025ProgressiveRiceSeed}. However, the reliance on handcrafted features limited the adaptability of these methods under complex and variable field conditions. Consequently, deep learning has been widely adopted, with architectures evolving from convolutional neural networks (CNN) to transformer-based and hybrid CNN–transformer models, enabling more expressive feature representations and substantially improving disease recognition accuracy~\citep{Zhang2025, Salihu2025, Ghosh2025, Islam2025, Zeng2025}. While accuracy continues to improve, the associated computational and memory demands remain a major barrier to deployment on resource-constrained agricultural devices.

To address these computational constraints, researchers have developed compact architectures by integrating lightweight channel-attention modules, such as squeeze-and-excitation mechanisms, into efficient backbones \citep{Lv2025, Liang2025, Duhan2025a}. Nevertheless, the compact design of these models can limit their ability to capture complex visual patterns, leading to reduced performance on fine-grained tasks. Knowledge distillation (KD) overcomes this trade-off by transferring discriminative knowledge from high-capacity teacher networks to small student models. Recent studies in smart agriculture have further enhanced KD through attention-based feature refinement, explainability-focused evaluation of feature alignment, and heterogeneous frameworks combining CNNs with Transformers to capture multi-scale semantic features~\citep{Dong2024, Duhan2025b, Quach2026, Li2026}.

However, despite these advancements, current KD frameworks in agriculture face three key challenges. First, student models are often still relatively large. For example, the student model proposed in~\citep{Quach2026} remains unsuitable for deployment on resource-constrained edge devices. Second, knowledge is often distilled only from the final layer (logits) using Kullback-Leibler (KL) divergence, limiting the transfer of rich information from the teacher~\citep{Dong2024, Duhan2025b, Quach2026}. Third, most studies evaluate their methods on a single dataset, which does not demonstrate the generalization of the models across diverse agricultural scenarios~\citep{Ding2024, Demirel2025, Quach2026}.

Motivated by the limitations of current knowledge distillation methods in smart agriculture, this study proposes a hybrid knowledge distillation framework that integrates a lightweight student model with multi-objective distillation, designed for resource-constrained devices. Specifically, we:

\begin{itemize}
    \item Designing a lightweight student architecture that combines inverted residual blocks, inspired by MobileNetV2, with DenseNet-style dense connectivity. This model maintains strong feature representation while being computationally efficient for edge deployment.
    
    \item Developing a hybrid knowledge distillation framework to transfer knowledge from a ResNet18 teacher to the student using a multi-objective strategy:
    \begin{itemize}
        \item \textit{Hard label supervision} ensures the student learns the main classification task.
        \item \textit{Feature-based distillation} guides the student to match the teacher's internal feature representations.
        \item \textit{Response-based distillation} teaches the student to mimic the teacher's output predictions, capturing complex inter-class relationships.
        \item \textit{Self-distillation} leverages auxiliary branches within the student to improve stability and generalization.
    \end{itemize}
    
    \item Evaluating performance across five agricultural datasets, including an author-constructed rice variety dataset and four public plant leaf disease datasets (rice, potato, coffee, and corn), to validate effectiveness and generalization.
    
    \item Applying Grad-CAM visualization to interpret the model's attention and highlight which regions contribute most to predictions, allowing comparison with other popular pretrained models.
\end{itemize}

The remainder of this paper is organized as follows. Section~\ref{sec:related_works} reviews related works in smart agriculture and knowledge distillation. Section~\ref{sec:methodology} introduces the proposed hybrid knowledge distillation framework and the lightweight student architecture. Section~\ref{sec:experiments} presents the experimental setup and datasets, while Section~\ref{sec:results} reports and analyzes the results. Finally, Section~\ref{sec:discussion} discusses the findings, and Section~\ref{sec:conclusion} concludes the paper with key contributions and future directions.

\section{Related works}\label{sec:related_works}

In conventional agricultural image analysis, research efforts have largely relied on traditional machine learning approaches built on handcrafted visual features. These approaches have been widely applied to agricultural image classification, with feature extraction focusing on manually designed color and texture representations. For instance,~\cite{Ahmed2021} applied region-of-interest segmentation before extracting Gray-Level Co-occurrence Matrix (GLCM) and color features, achieving 98.79\% accuracy on the PlantVillage dataset using a Support Vector Machine (SVM). Similarly,~\cite{Rusli2022} proposed a potato leaf disease classification method in which diseased regions were segmented using the K-means algorithm, texture features were extracted with GLCM, and an artificial neural network (ANN) was employed for classification, achieving 94\% accuracy. To enhance discriminative power,~\cite{Alsakar2024} introduced a feature fusion framework based on color correlograms and multi-level multi-channel Local Binary Patterns (LBP) extracted from image blocks, achieving 99.53\% accuracy for rice leaf disease classification. In the context of rice seed variety recognition, traditional machine learning approaches based on handcrafted visual features have also been extensively explored. \cite{Phan2024RiceSeedHandcrafted} investigated rice seed purity classification using a combination of handcrafted visual descriptors, including GLCM, Scale-Invariant Feature Transform (SIFT), Histogram of Oriented Gradients (HOG), and GIST features, achieving an accuracy of 94.73\%. Building upon similar feature representations, \cite{Le2025ProgressiveRiceSeed} proposed a progressive feature enrichment strategy that systematically integrates morphological, color, and texture features, coupled with ensemble learning, improving the average classification accuracy to 97.58\% on the same rice seed dataset. 
Although these handcrafted-feature-based methods demonstrate competitive performance, they rely heavily on manual feature design and task-specific tuning, which limits scalability and generalization in complex real-world agricultural scenarios. This limitation motivates the adoption of deep learning approaches that can automatically learn hierarchical and discriminative representations directly from data.

Deep learning approaches have been widely adopted in agricultural image analysis to overcome the limitations of handcrafted feature-based methods. Prior studies have demonstrated the effectiveness of standard convolutional neural networks (CNN), such as the customized CNN architecture proposed by~\cite{Salihu2025} for potato leaf disease detection that achieved 96.88\% accuracy, and the CNN-based framework introduced by~\cite{Dame2025} for potato leaf disease detection and severity assessment, evaluated against classical architectures including AlexNet and VGG16, with an overall disease classification accuracy of 99\% and an average detection accuracy of 96\%. To further improve performance, subsequent research has investigated more complex hybrid architectures.~\cite{Ghosh2025} combined Swin transformer, Vision transformer (ViT), and EfficientNetV2 within a unified framework for tomato leaf disease classification and reported an accuracy of 98\%. Beyond plant disease classification, fine-grained recognition problems in agriculture have also been investigated.~\cite{Zhang2025} introduced RSCD-Net, a deep residual architecture that integrates a space and channel feature extraction residual block (SCR-Block) with a double attention mechanism (A$^{2}$Net). The model was specifically designed to improve inter-class differentiation, reduce redundant features, and capture subtle visual variations among rice varieties through joint spatial–channel feature extraction and double attention. In parallel, lightweight architectures have been developed to improve computational efficiency. For instance,~\cite{Liang2025} enhanced the EfficientNet Lite0 model with a channel focus attention (CFA) mechanism, a feature reuse module (FRM), and a feature enhancement mobile inverted bottleneck convolution module (FEMBCM), achieving 98.03\% accuracy for tea disease classification under complex backgrounds. Similarly,~\cite{Duhan2025a} integrated various attention mechanisms into the MobileNetV2 framework, reporting accuracies ranging from 82\% to 100\% across multiple plant leaf disease datasets while significantly reducing its computational footprint. Although high recognition accuracy has been reported, the growing architectural complexity of these models, particularly hybrid and attention-enhanced networks, poses substantial challenges for deployment on edge devices with limited computational resources.

To bridge the gap between high recognition performance and deployment on resource-constrained edge devices, knowledge distillation has recently gained attention in smart agriculture~\citep{Huang2023, Sun2026LigTomDet}. This approach transfers knowledge from a high-capacity teacher to a smaller student network, which allows the student to achieve high accuracy with reduced computational cost. Recent research has further optimized this process through various strategies, ranging from advanced distillation losses to the integration of attention mechanisms. For instance,~\cite{Ding2024} enhanced a ShuffleNetV2 student by applying masked generative distillation to transfer feature-level spatial details, achieving an F1-score of 91.92\%. Integrating attention mechanisms,~\cite{Dong2024} trained an ECA-enhanced EfficientNet-B0 student via Kullback-Leibler divergence loss from a ConvNeXt teacher to reach 98.28\% accuracy, while~\cite{Duhan2025b} optimized the ultra-lightweight LiteCShuffle student with Channel Attention Module (CAM) using standard logits distillation, attaining 99.86\% accuracy with only 0.15 million parameters. Furthermore,~\cite{Demirel2025} proposed the GGENet student with global response normalization, utilizing adaptive knowledge distillation to transfer attention maps, resulting in 98.18\% accuracy. Moving beyond architecture optimization to interpretability,~\cite{Quach2026} employed a teacher-student framework to guide a ResNet50 student, utilizing Grad-CAM combined with Intersection over Union (IoU) metrics to quantify the alignment between the model's attention regions and the actual pest locations. More recently,~\cite{Li2026} proposed a hybrid teacher combining ResNet50 for local feature extraction and PVTv2 (Pyramid Vision Transformer) for global context modeling. Knowledge was distilled to a lightweight student built with weight sharing dilated partial convolutions (WSD-PConv), which reduces parameter count and memory access redundancy via partial channel processing and kernel sharing, while enabling multi-scale feature extraction and FLOPs-efficient low-latency inference. A staircase multi-branch distillation strategy with Wasserstein distance regularization Kullback-Leibler (WDRKL) loss guided the student, achieving accuracies ranging from 96.86\% to 98.67\% across multiple plant leaf disease datasets.

\section{Methodology}\label{sec:methodology}

\subsection{Overview of the proposed approach}

\begin{figure}[H]
    \hspace{-1.2cm}
    \includegraphics[width=1.2\textwidth]{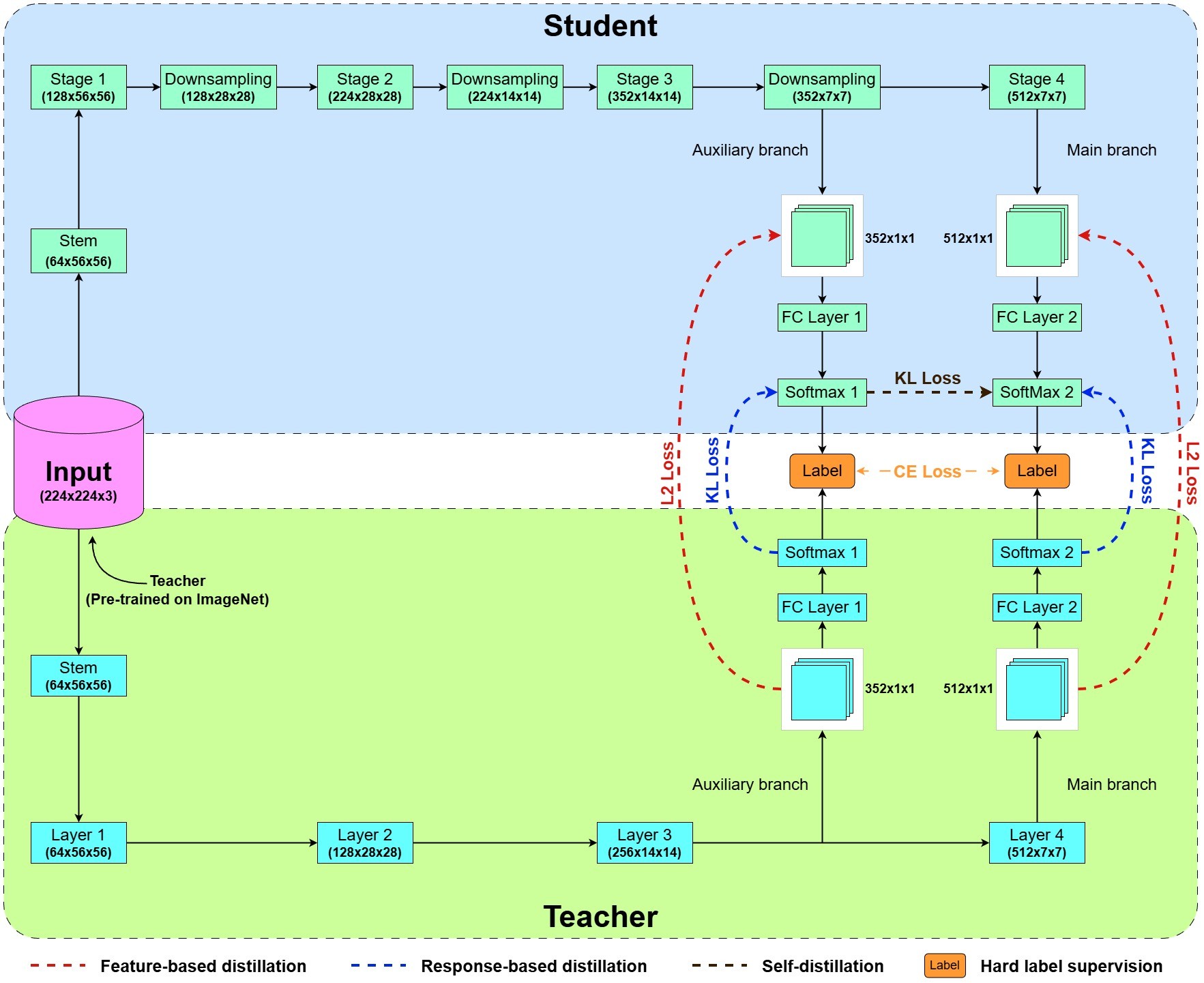}
    \caption{Overall framework of the proposed knowledge distillation approach}
    \label{fig:overall_framework}
\end{figure}

In this study, we propose a novel hybrid knowledge distillation (KD) framework aimed at developing a lightweight model for the accurate classification of rice varieties and leaf diseases on resource-constrained devices in real-world agricultural settings. The overall pipeline of the framework is shown in Figure~\ref{fig:overall_framework}.
The framework comprises three main components as follows:

\vspace{0.2cm}

1. Novel lightweight student architecture  

\vspace{0.2cm}

The student model adopts a hybrid design that integrates MobileNetV2-style inverted residual blocks - consisting of linear bottlenecks and depthwise separable convolutions - with DenseNet-style dense connectivity. Furthermore, the computationally expensive $1\times1$ transition convolutions typically used in DenseNet are completely removed to further reduce the
computational cost. This design choice preserves the efficiency of MobileNetV2-style operations, enhances feature reuse through dense connectivity, and results in a more compact architecture with fewer parameters and FLOPs compared with existing MobileNetV2–DenseNet hybrid variants.

\vspace{0.2cm}

2. Teacher model  
\vspace{0.2cm}

   ResNet18 is employed as the teacher. Its residual connections support stable optimization and robust feature extraction, while its moderate depth provides rich hierarchical representations without creating an excessively large capability gap relative to the lightweight student. This balanced complexity makes ResNet18 well suited for guiding the student model in an online distillation setting.

\vspace{0.2cm}
3.  Hybrid knowledge distillation strategy with multi-objective optimization
\vspace{0.2cm}

   The knowledge transfer process employs both a main classification branch, which provides primary supervision, and an auxiliary branch that aligns intermediate feature representations with the teacher. A hybrid distillation strategy is applied, combining four complementary loss components:  
   
   (i) hard-label supervision using cross-entropy,  
   
   (ii) feature-based distillation via L2 alignment at Stages 3 and 4,
   
   (iii) response-based distillation using KL divergence, and  
   
   (iv) self-distillation to encourage consistency between the student's two branches.  
   
   These loss terms jointly form a multi-objective optimization scheme that stabilizes training and enables effective transfer of both low-level feature representations and high-level relational knowledge from the teacher.

The following section presents a detailed description of the methods used in this study.

\subsection{Design of the novel lightweight student architecture}

A novel lightweight Convolutional Neural Network (CNN) architecture is proposed as the student model, designed to achieve a strong balance between computational efficiency and expressive feature representation. The architecture draws inspiration from the inverted residual structure of MobileNetV2 and the dense connectivity mechanism of DenseNet, allowing the model to maintain high representational capacity while remaining suitable for deployment on resource-constrained edge devices.

\subsubsection{Macro-architecture}

The overall structural layout of the proposed student network is presented in Table~\ref{tab:overall_student_architecture}. Designed with the objective of maximizing feature expressiveness under a constrained computational cost, the architecture follows a four-stage hierarchical organization that progressively transforms low-level visual information into increasingly abstract and discriminative representations.

\begin{table}[H]
\centering
\caption{Overall architecture of the student model with parameter details.}
\renewcommand{\arraystretch}{1.4}
\setlength{\tabcolsep}{3pt}
\resizebox{\columnwidth}{!}{
\begin{tabular}{|l|c|c|c|}
\hline
\textbf{Layers} & \textbf{Output Size} & \textbf{Student} & \textbf{Params} \\
\hline
\multirow{2}{*}{Stem} & 64$\times$112$\times$112 & 7$\times$7 conv, stride 2 & \multirow{2}{*}{0.01M} \\
\cline{2-3}
 & 64$\times$56$\times$56 & 3$\times$3 max pool, stride 2 & \\
\hline
Stage 1 & 128$\times$56$\times$56 &
  $\left[ \begin{tabular}{@{}c@{}} 1$\times$1 conv, stride 1 \\ inverted residual layer \end{tabular} \right] \times 4$ & 0.1M \\
\hline
Downsampling layer 1 & 128$\times$28$\times$28 & 2$\times$2 average pool, stride 2 & 0 \\
\hline
Stage 2 & 224$\times$28$\times$28 &
  $\left[ \begin{tabular}{@{}c@{}} 1$\times$1 conv, stride 1 \\ inverted residual layer \end{tabular} \right] \times 6$ & 0.18M \\
\hline
Downsampling layer 2 & 224$\times$14$\times$14 & 2$\times$2 average pool, stride 2 & 0 \\
\hline
Stage 3 & 352$\times$14$\times$14 &
  $\left[ \begin{tabular}{@{}c@{}} 1$\times$1 conv, stride 1 \\ inverted residual layer \end{tabular} \right] \times 8$ & 0.3M \\
\hline
Downsampling layer 3 & 352$\times$7$\times$7 & 2$\times$2 average pool, stride 2 & 0 \\
\hline
Stage 4 & 512$\times$7$\times$7 &
  $\left[ \begin{tabular}{@{}c@{}} 1$\times$1 conv, stride 1 \\ inverted residual layer \end{tabular} \right] \times 10$ & 0.47M \\
\hline
\multirow{2}{*}{Classification layer} & 512$\times$1$\times$1 & 7$\times$7 global average pool & \multirow{2}{*}{512$\times C$M} \\ 
\cline{2-3}
 & & $C$-D fully-connected, softmax & \\
\hline
\multicolumn{3}{|c|}{\textbf{Total parameters}} & \textbf{$(1.06 + 512C)$M} \\
\hline
\multicolumn{4}{l}{\footnotesize{Note: $C$ denotes the number of classes; M denotes millions of parameters.}} \\
\end{tabular}
}
\label{tab:overall_student_architecture}
\end{table}

To facilitate effective feature extraction from the earliest stages, the architecture employs a ``\textit{Stem}'' design similar to established models like ResNet and DenseNet. Specifically, the network begins with a $7\times7$ convolutional layer (64 output channels) followed by a $3\times3$ max pooling layer, both with a stride of 2. This configuration quickly downsamples the input spatial dimensions, effectively preserving essential low-level structural information while conserving computational resources for the subsequent layers.

Following the initial feature extraction and dimensionality reduction in the Stem, the network body is organized into four hierarchical stages (stage 1 to stage 4). As detailed in Table~\ref{tab:overall_student_architecture}, each stage consists of multiple blocks, which combine a $1\times1$ convolution (DenseNet-B bottleneck layer, see~\cite{Huang2018}) and an inverted residual structure with a linear shortcut. The network depth increases progressively across the four stages, with each stage containing 4, 6, 8, and 10 blocks, respectively. This gradual increase in the number of blocks allows the network to learn increasingly complex and abstract feature representations, effectively capturing both low-level and high-level semantic information as data flows through the stages.

To effectively manage spatial resolution and computational complexity, explicit downsampling layers are inserted between stages. These layers play a role analogous to the transition layers in DenseNet, however, to further reduce computational overhead, the $1\times1$ convolution typically used in transition blocks is omitted, retaining only a $2\times2$ average pooling operation (stride = 2). As a result, the spatial resolution is reduced by half at each transition, from $56\times56$ after Stage~1, to $28\times28$, $14\times14$, and ultimately $7\times7$ after Stage~4, while the channel depth increases from 128 to 224, 352, and 512. This progressive reduction in feature-map size, combined with controlled expansion of channel dimensionality, enables the model to capture increasingly abstract semantic information without inflating the parameter count.

Finally, the network is completed with a classification layer. The $512\times7\times7$ feature maps produced by stage 4 are aggregated using a $7\times7$ global average pooling (GAP) operation to obtain a compact $512\times1\times1$ feature vector. This vector is subsequently passed through a fully connected layer followed by a Softmax activation to generate the final class probability distribution.

\subsubsection{Micro-architecture}

\begin{figure}[H]
    \centering
    \includegraphics[width=1.0\textwidth]{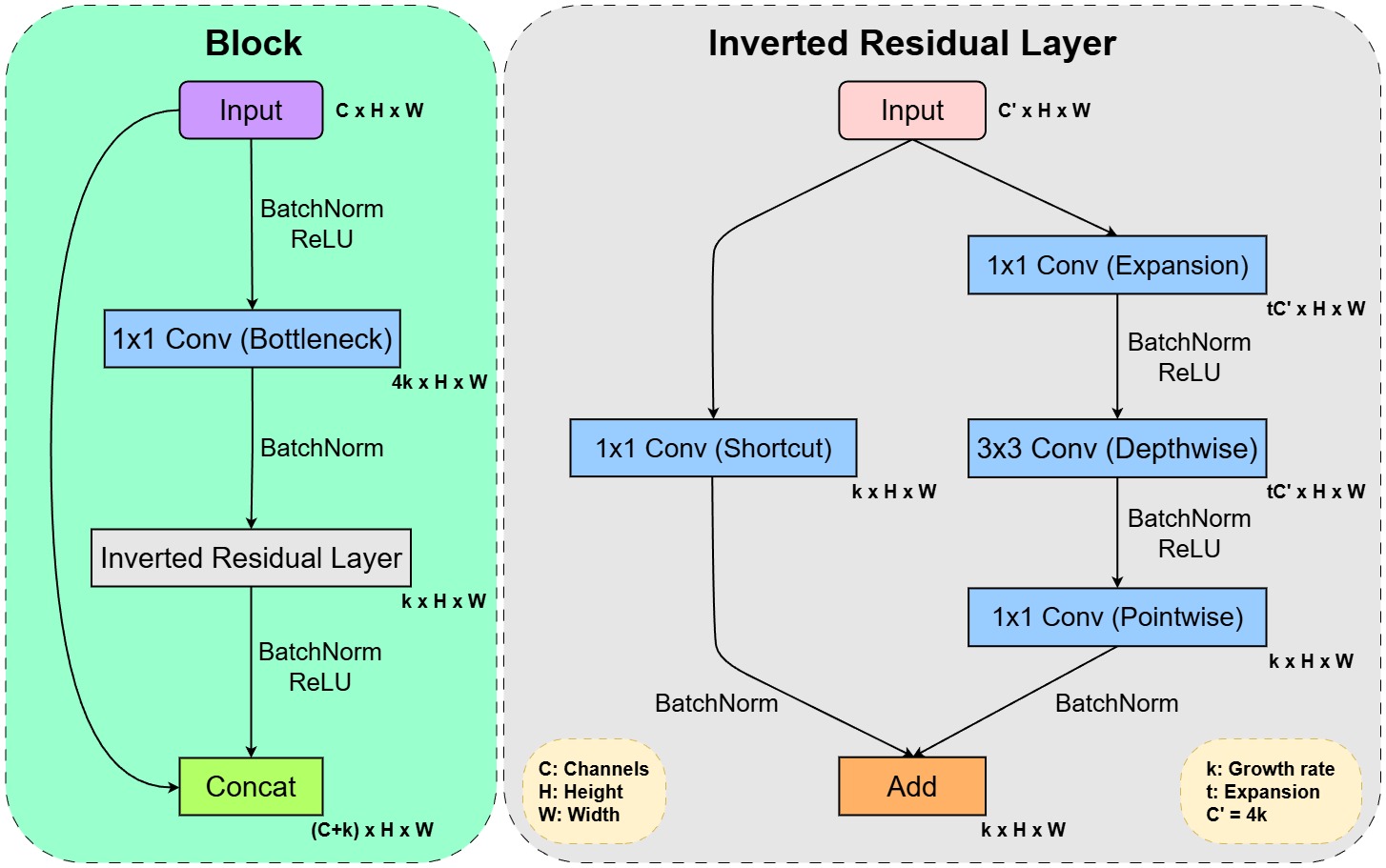}
    \caption{Architecture of a block in the student model.}
    \label{fig:student_block_architecture}
\end{figure}

While the macro-architecture outlines the hierarchical organization of the network, the micro-architecture defines the specific design of the fundamental building block. Each block is constructed as a hybrid structure that combines the feature reuse mechanism of DenseNet~\citep{Huang2018} with the efficiency of the MobileNetV2~\citep{Mark2019} inverted residual design. As illustrated in Figure~\ref{fig:student_block_architecture}, the block consists of two main components: the \textit{bottleneck layer} and the \textit{inverted residual layer}.

\textbf{a) Bottleneck layer}

\vspace{0.2cm}

The block begins with a $1\times1$ convolution applied to the input feature map of size $C \times H \times W$. Following the DenseNet-B bottleneck design, this layer reduces the input dimensionality before further processing. A compact growth rate of $k=16$ is adopted, projecting the features to $4k$ channels and producing an intermediate representation $\mathbf{x}_{\text{bot}} \in \mathbb{R}^{4k \times H \times W}$. This design maintains the representational benefits of the bottleneck while keeping the computational cost low, which is essential for lightweight architectures.

\vspace{0.2cm}

\textbf{b) Inverted residual layer}

\vspace{0.2cm}

The inverted residual layer processes $\mathbf{x}_{\text{bot}}$ through two parallel branches that are merged by addition. In the main branch, a $1\times1$ expansion convolution increases the feature dimensionality, followed by a $3\times3$ depthwise convolution and a $1\times1$ pointwise projection that returns the features to $k$ channels. The expansion ratio is set to $t=3$ to balance capacity and efficiency.

In parallel, the shortcut branch replaces the identity mapping of conventional residual blocks with a linear $1\times1$ projection to ensure dimensional compatibility. The final output of the block is formulated as:

\begin{equation}
\mathbf{x}_{\text{res}} = F(\mathbf{x}_{\text{bot}}) + W(\mathbf{x}_{\text{bot}}) \in \mathbb{R}^{k \times H \times W},
\label{eq:output_inverted_residual_layer}
\end{equation}

where $F(\cdot)$ denotes the nonlinear transformation in the main branch and $W(\cdot)$ represents the linear shortcut projection.

\textbf{c)} Dense connectivity

Finally, the block employs a dense connectivity mechanism to further enhance feature reuse and information flow across layers. The input is concatenated with the features produced by the bottleneck and inverted residual layers. Let $x_{\text{in}} \in \mathbb{R}^{C \times H \times W}$ denote the input of the block and $x_{\text{res}} \in \mathbb{R}^{k \times H \times W}$ the output of the inverted residual layer, where $k$ is the growth rate. The final block output is then formulated as:

\begin{equation}
y = [x_{\text{in}}, x_{\text{res}}] \in \mathbb{R}^{(C+k) \times H \times W}
\label{eq:output_block}
\end{equation}

\subsection{Teacher architecture}

In knowledge distillation, the teacher model serves as the guiding reference, and its capability directly determines the upper bound of the student model's performance. An effective teacher must be able to capture complex patterns within the data in order to provide high-quality supervisory signals. For this reason, selecting an appropriate teacher architecture is a foundational step. Following this consideration, ResNet18 is adopted as the teacher model in our framework due to its demonstrated effectiveness and strong suitability for distillation tasks. In our implementation, the teacher network is initialized from ImageNet-pretrained weights to strengthen the quality of its soft-target supervision. Importantly, the teacher is not frozen; it is jointly optimized with the student throughout training under the online distillation setting.

First, the effectiveness of ResNet architectures as teacher models has been consistently supported by empirical evidence. Several studies, including~\cite{Zagoruyko2017},~\cite{Park2019}, and~\cite{Gou2021}, have demonstrated that the ResNet family provides stable and expressive feature representations that are crucial for enhancing the performance of student networks in knowledge distillation.

Second, among the ResNet variants, ResNet18 provides a well-balanced and practical design. It offers sufficient depth to learn rich and discriminative features while avoiding the heavy computational cost and diminishing returns seen in deeper models such as ResNet34 or ResNet50. Furthermore, the residual shortcut connections characteristic of the ResNet architecture help maintain stable optimization and support the learning of deep and informative representations. These properties enable the teacher model to capture robust patterns that form strong supervisory signals for the student.

\subsection{ Design of an auxiliary branch for intermediate feature distillation}

\begin{figure}[H]
    \centering
    \includegraphics[width=1.0\textwidth]{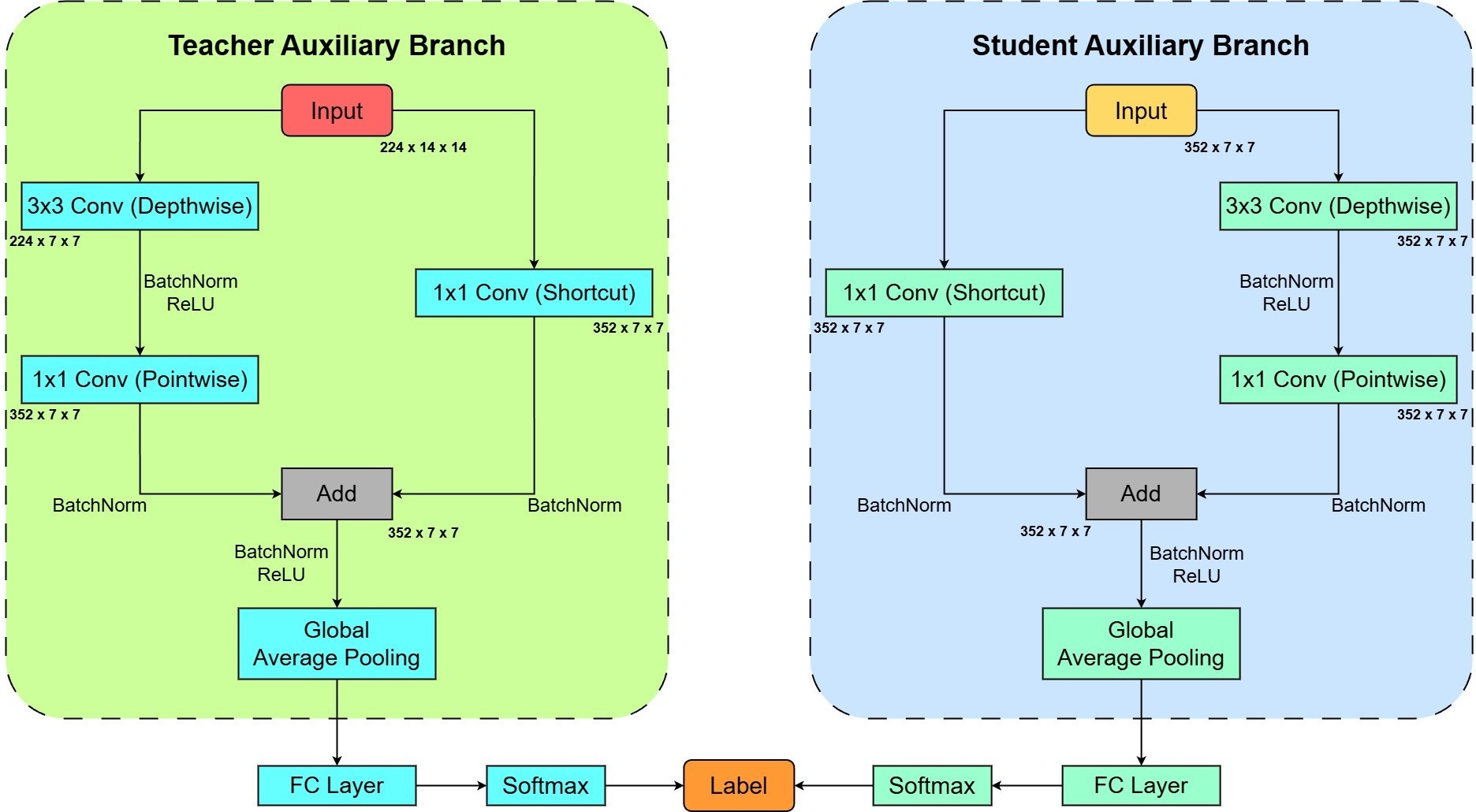}
    \caption{Design of the auxiliary branches used in the teacher and student models}
    \label{fig:teacher_student_auxiliary_branch}
\end{figure}

To strengthen knowledge transfer, we incorporate an intermediate distillation path from the teacher's layer-3 output to the student's third downsampling block. Intermediate supervision enables the student to capture the teacher's representational hierarchy, rather than only its final predictions, allowing it to learn both fine-grained features and higher-level semantic patterns. However, the feature maps at these layers differ in both spatial resolution and channel depth, as the teacher outputs a $224 \times 14 \times 14$ tensor while the student produces $352 \times 7 \times 7$. This mismatch prevents direct feature alignment. To address this, we introduce auxiliary alignment branches for both networks, as illustrated in Figure~\ref{fig:teacher_student_auxiliary_branch}. A key design choice is to expand the teacher's channels from 224 to 352, rather than compressing the student's representation, thereby avoiding information loss and ensuring that the aligned teacher features occupy the same representational space as the student's.

Both the teacher and student auxiliary branches employ an Inverted Residual structure, which decomposes standard convolutions into depthwise and pointwise operations to preserve computational efficiency while enabling flexible feature transformation. In the teacher branch, the main path applies a $3 \times 3$ depthwise convolution with stride 2, followed by a $1 \times 1$ pointwise convolution that expands the channel depth to 352. A parallel $1 \times 1$ shortcut projection is added to the main path output to produce the final aligned feature map and stabilize gradient flow.

The student auxiliary branch mirrors this architecture but uses a stride of 1, preserving the input resolution of $352 \times 7 \times 7$. Although the student's third downsampling layer already matches the teacher's feature dimensions, the Inverted Residual block is retained to maintain architectural consistency and to refine the student's representation so that it captures a comparable level of semantic detail. Finally, the outputs from both auxiliary branches are passed through a global average pooling layer to generate compact feature vectors for intermediate distillation.

\subsection{Construction of the hybrid distillation loss}

To effectively transfer knowledge from a high-capacity teacher model to a lightweight student model, this study adopts a sequential online knowledge distillation strategy. In this formulation, the teacher and student are trained within a single training process, while their parameters are updated sequentially at each mini-batch. Specifically, the teacher is first optimized using ground-truth supervision and then immediately provides updated feature representations and output distributions to guide the student within the same training iteration. Importantly, although the teacher is initialized with ImageNet-pretrained weights, it is not frozen; it continues to adapt to the target domain during co-training.

This choice is motivated by the limitations of offline distillation, where the teacher must first be fully trained and fixed before guiding the student. Such a two-stage process substantially increases computational cost and prevents the teacher from adapting to dataset-specific characteristics, an important factor in agricultural imagery, which exhibits strong domain variations. Prior studies on online distillation~\citep{Zhang2017, Guo2020, Zhang2021} have also shown that co-evolving teachers provide smoother and more learnable soft targets, particularly beneficial for compact students and for reducing the capacity-gap problem.

Under this sequential online distillation framework, the student learns not only from the teacher's feature representations and output distributions but also from its own intermediate representations through self-distillation, resulting in improved training stability and generalization. Building upon conventional knowledge distillation objectives, we construct an extended hybrid loss by integrating feature-level alignment and self-distillation into an online distillation framework.

\subsubsection{Teacher network's loss function}

In the proposed framework, the teacher network is optimized to learn independently and provide stable guidance for the student. Its optimization relies solely on the ground-truth labels, without any influence from the student. The teacher network is structured with a main branch and an auxiliary branch to capture representations at multiple levels of depth. 

The teacher's total optimization objective, denoted as $\mathcal{L}_{\text{Teacher}}$, is therefore formulated as the direct sum of the standard cross-entropy (CE) losses from both branches:

\begin{equation}
    \mathcal{L}_{\text{Teacher}} = \mathcal{L}_{\text{CE}}(T_{main}, y) + \mathcal{L}_{\text{CE}}(T_{aux}, y)
    \label{eq:teacher_loss}
\end{equation}

where $T_{main}$ and $T_{aux}$ are the output logits of the main and auxiliary branches, respectively, and $y$ is the ground-truth label. This multi-branch supervision encourages the teacher to learn more robust and informative representations, which can subsequently support more effective knowledge transfer to the student.

\subsubsection{Student network's loss function}

The student network is optimized using a composite loss function that integrates direct supervision from ground-truth labels with multiple knowledge distillation components. Each part of this objective represents a distinct pathway for transferring knowledge, as illustrated in the overall framework in Figure~\ref{fig:overall_framework}.

\textbf{a)} Hard label supervision

The foundational component of the student's learning process is direct supervision from the ground-truth labels. This ensures that the student network is grounded in empirical data, learning to perform the classification task accurately on its own rather than solely imitating the teacher. Following the same deep supervision principle as the teacher, the student is equipped with both a main and an auxiliary branch. This multi-branch architecture supports robust feature learning at different semantic levels and facilitates stable gradient flow during training. 

The hard label supervision objective, denoted as $\mathcal{L}_{\text{Hard}}$, is defined as the sum of the standard cross-entropy (CE) losses from both branches:

\begin{equation}
    \mathcal{L}_{\text{Hard}} = \mathcal{L}_{\text{CE}}(S_{main}, y) + \mathcal{L}_{\text{CE}}(S_{aux}, y)
    \label{eq:cross_entropy_loss}
\end{equation}

where $S_{main}$ and $S_{aux}$ are the output logits from the student's main and auxiliary branches, respectively, and $y$ is the ground-truth label.

\textbf{b)} Feature-based distillation

To ensure the student learns not only what the teacher predicts but also how it forms its representations, we incorporate feature-based distillation. This component encourages the student to align its internal representations with those of the teacher, effectively mimicking the teacher's feature extraction process. Directly minimizing the distance between the full feature maps of corresponding layers would be computationally expensive due to their high dimensionality. To address this, we adopt a more efficient approach by distilling knowledge from the feature vectors produced immediately after the \texttt{global average pooling (GAP)} layer in both the main and auxiliary branches. This design provides two key advantages:

\begin{itemize}
    \item \textbf{Computational efficiency:} By comparing the compact feature vectors after \texttt{GAP} rather than the dense feature maps, the computational cost is significantly reduced, making training more efficient.

    \item \textbf{Channel-wise attention guidance:} This approach emphasizes the most important feature channels identified by the teacher, rather than enforcing strict spatial alignment. As a result, the student learns to focus on the same salient information as the teacher, capturing a summary of the teacher's attention across feature channels.
    
\end{itemize}

The feature-based distillation loss, denoted as $\mathcal{L}_{\text{FD}}$, is defined as the Euclidean distance between the corresponding feature vectors:

\begin{equation}
    \mathcal{L}_{\text{FD}} = \left\| f_{S_{main}} - f_{T_{main}} \right\|_2 + \left\| f_{S_{aux}} - f_{T_{aux}} \right\|_2
    \label{eq:feature_based_distillation_loss}
\end{equation}

where $f_{S_{main}}$ and $f_{T_{main}}$ are the feature vectors from the main branches of the student and teacher after \texttt{GAP}, and $f_{S_{aux}}$ and $f_{T_{aux}}$ are the corresponding vectors from the auxiliary branches.

\textbf{c)} Response-based distillation

Beyond aligning internal features, response-based distillation transfers the teacher's high-level, class-level knowledge by focusing on its final output distributions~\citep{Hinton2015}. This component trains the student to mimic the class probabilities produced by the teacher, often called \emph{soft targets}. Unlike one-hot labels, these soft targets encode both inter-class similarities and the teacher's confidence. If the teacher's output assigns a high probability to only a single class, the student cannot learn the relationships among classes. By distributing probability across multiple classes, soft targets provide richer relational information that allows the student to capture similarities and distinctions between classes that ground-truth labels alone cannot convey.

The difference between the student's and teacher's softened outputs is measured using the Kullback-Leibler (KL) divergence, which quantifies how one probability distribution differs from another. To capture these relationships, we apply a temperature parameter $\tau$ in the softmax function. Increasing $\tau$ softens the output probabilities, distributing them more evenly across classes, which allows the student to learn not only the correct class but also the similarities and relationships among classes. Decreasing $\tau$ sharpens the distribution, concentrating probability on the most likely class and emphasizing the dominant prediction.

The response-based distillation loss, denoted as $\mathcal{L}_{\text{RD}}$, is defined as the sum of the Kullback-Leibler (KL) divergence between the softened output distributions of the student and teacher for both branches:

\begin{equation}
    \mathcal{L}_{\text{RD}} = \mathcal{L}_{\text{KL}}(\sigma(S_{main}/\tau), \sigma(T_{main}/\tau)) + \mathcal{L}_{\text{KL}}(\sigma(S_{aux}/\tau), \sigma(T_{aux}/\tau))
    \label{eq:responsed_based_distillation_loss}
\end{equation}

where $S_{main}$, $T_{main}$, $S_{aux}$, and $T_{aux}$ are the logits from the respective branches of the student and teacher, $\sigma(\cdot)$ is the softmax function, and $\tau$ is the temperature hyperparameter.

\textbf{d)} Self-distillation

Inspired by self-refinement strategies, we propose a self-distillation mechanism to support intra-network knowledge transfer. While teacher–student distillation relies on an external expert, the student model also contains internal layers of information that can be used to improve generalization. In this approach, the auxiliary branch acts as an internal mentor for the main branch.

The main branch is deeper and capable of learning higher-level abstractions, yet it is more vulnerable to overfitting and gradient degradation in the early training stages. The auxiliary branch, derived from intermediate layers that are closer to the input, converges more rapidly and captures stable structural cues and generalized patterns.

By using the outputs of the auxiliary branch to guide the main branch, deeper layers are encouraged to form richer abstractions while remaining anchored to the foundational visual evidence detected by the shallower layers. Consistency between the two branches acts as a regularizer, ensuring that the final prediction integrates both high-level semantics and core feature representations. This process consolidates the model's internal knowledge into a coherent progression from low-level feature extraction to high-level decision-making.

The self-distillation loss, denoted as $\mathcal{L}_{\text{SD}}$, is formulated using the Kullback-Leibler (KL) divergence between the softened output distributions of the student's two branches:

\begin{equation}
    \mathcal{L}_{\text{SD}} = \mathcal{L}_{\text{KL}}(\sigma(S_{main} / \tau'), \sigma(S_{aux} / \tau'))
    \label{eq:self_distillation_loss}
\end{equation}

where $S_{main}$ and $S_{aux}$ are the logits from the student's main and auxiliary branches, $\sigma(\cdot)$ is the softmax function, and $\tau'$ is a temperature hyperparameter that smooths the probability distributions to enable effective internal knowledge transfer.

\textbf{e)} Overall loss function for student

The student network is optimized using a composite loss that integrates direct supervision with the multiple knowledge transfer mechanisms described above. The total loss, denoted as $\mathcal{L}_{\text{Student}}$, consists of four components: hard label supervision (\(\mathcal{L}_{\text{Hard}}\)), feature-based distillation (\(\mathcal{L}_{\text{FD}}\)), response-based distillation (\(\mathcal{L}_{\text{RD}}\)), and self-distillation (\(\mathcal{L}_{\text{SD}}\)), combined into a weighted sum:

\begin{equation}
    \mathcal{L}_{\text{Student}} = \lambda_1 \mathcal{L}_{\text{Hard}} + \lambda_2 \mathcal{L}_{\text{FD}} + \lambda_3 \mathcal{L}_{\text{RD}} + \lambda_4 \mathcal{L}_{\text{SD}}
    \label{eq:student_loss}
\end{equation}

where $\lambda_1$, $\lambda_2$, $\lambda_3$, and $\lambda_4$ are weighting hyperparameters controlling the contribution of each term to the total loss with values selected from the range (0, 1).

\section{Experiments}\label{sec:experiments}

\subsection{Data description}

In this study, we use a total of five different datasets to evaluate the effectiveness of the proposed method. The primary dataset is a rice variety dataset that we have constructed and introduced. The remaining four datasets focus on leaf diseases in rice, coffee, corn, and potato. These datasets are employed to demonstrate the generalization capability and potential applicability of the model across different image recognition tasks.

\subsubsection{Rice variety dataset}

This study employs an image dataset comprising nine Vietnamese rice seed varieties, including BC-15, Huong Thom-1, Nep-87, Q-5, TBR-36, TBR-45, TH-35, Thien Uu-8, and Xi-23. Several of these varieties exhibit highly similar visual appearances in terms of seed shape, size, surface texture, and color. The discriminative morphological cues among varieties are often subtle and difficult to perceive without specialized expertise, which poses significant challenges for automated recognition.

To reflect practical seed inspection and purity verification scenarios, the dataset is organized following a one-versus-rest binary classification scheme, consistent with experimental settings adopted in previous studies on seed purity assessment~\citep{Phan2015Comparative, Phan2025EnhancingRiceSeed, Le2025ProgressiveRiceSeed}. For each rice variety, images of that variety are treated as positive samples representing authentic seeds, while images from the remaining eight varieties are collectively regarded as negative samples corresponding to contaminated or non-pure seeds. This formulation decomposes the task into independent binary decisions, each aiming to determine whether a given seed belongs to the target variety or not.

To mitigate bias caused by class imbalance, the numbers of positive and negative samples are carefully controlled, with the difference between the two classes kept within a relatively small margin. Table \ref{tab:Rice_dataset} summarizes the distribution of images for each rice variety. For example, the Xi-23 subset contains 2,340 positive images of Xi-23 seeds, while its 2,239 negative samples are randomly selected from the other eight varieties. This experimental design closely mimics real-world inspection conditions, where the primary objective is to detect impurity rather than to identify all seed types.

\begin{figure}[H]
    \centering
    \includegraphics[width=1.0\textwidth]{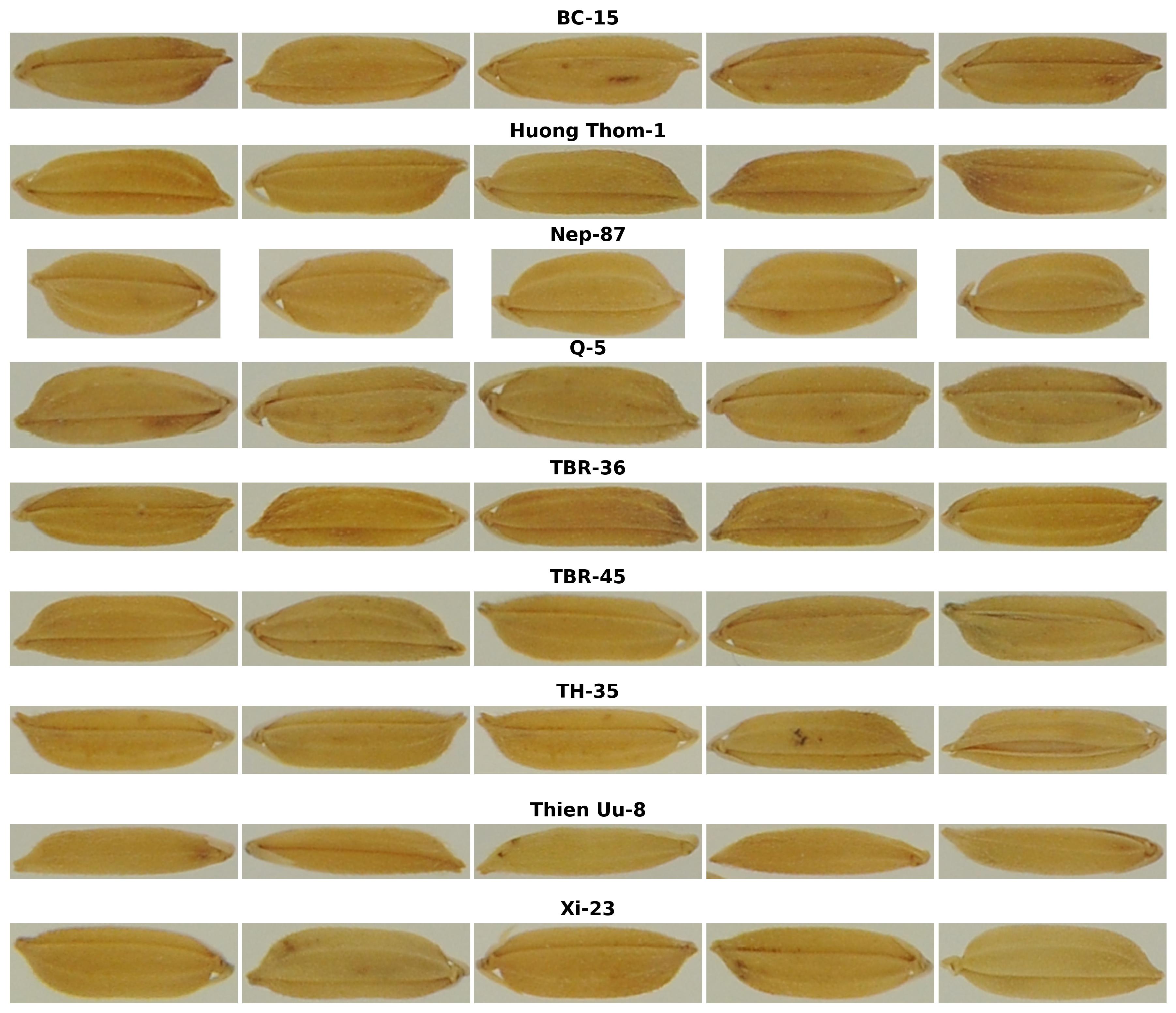}
    \caption{Sample images from the rice variety dataset containing 9 different rice varieties.}
    \label{fig:rice_seed_dataset}
\end{figure}

\begin{table}[H]
    \centering
    \footnotesize
    \caption{Description of 9 types of rice seed image dataset}
    \label{tab:Rice_dataset}
    \begin{tabular}{|l|c|c|c|}
        \hline
        \textbf{Rice seed name} & \textbf{Positive} 
        & \textbf{Negative} & \textbf{Total images} \\ \hline
        BC-15        & 1834 & 1925 & 3759 \\ \hline
        Huong Thom-1 & 2116 & 2200 & 4316 \\ \hline
        Nep-87       & 1399 & 1468 & 2867 \\ \hline
        Q-5          & 1924 & 2020 & 3944 \\ \hline
        TBR-36       & 1136 & 1192 & 2328 \\ \hline
        TBR-45       & 1140 & 1197 & 2337 \\ \hline
        TH-35        & 1012 & 1062 & 2074 \\ \hline
        Thien Uu-8   & 1026 & 1077 & 2103 \\ \hline
        Xi-23        & 2340 & 2239 & 4579 \\ \hline
    \end{tabular}
\end{table}

\subsubsection{Rice leaf disease dataset}

\begin{figure}[H]
    \centering
    \includegraphics[width=0.7\textwidth]{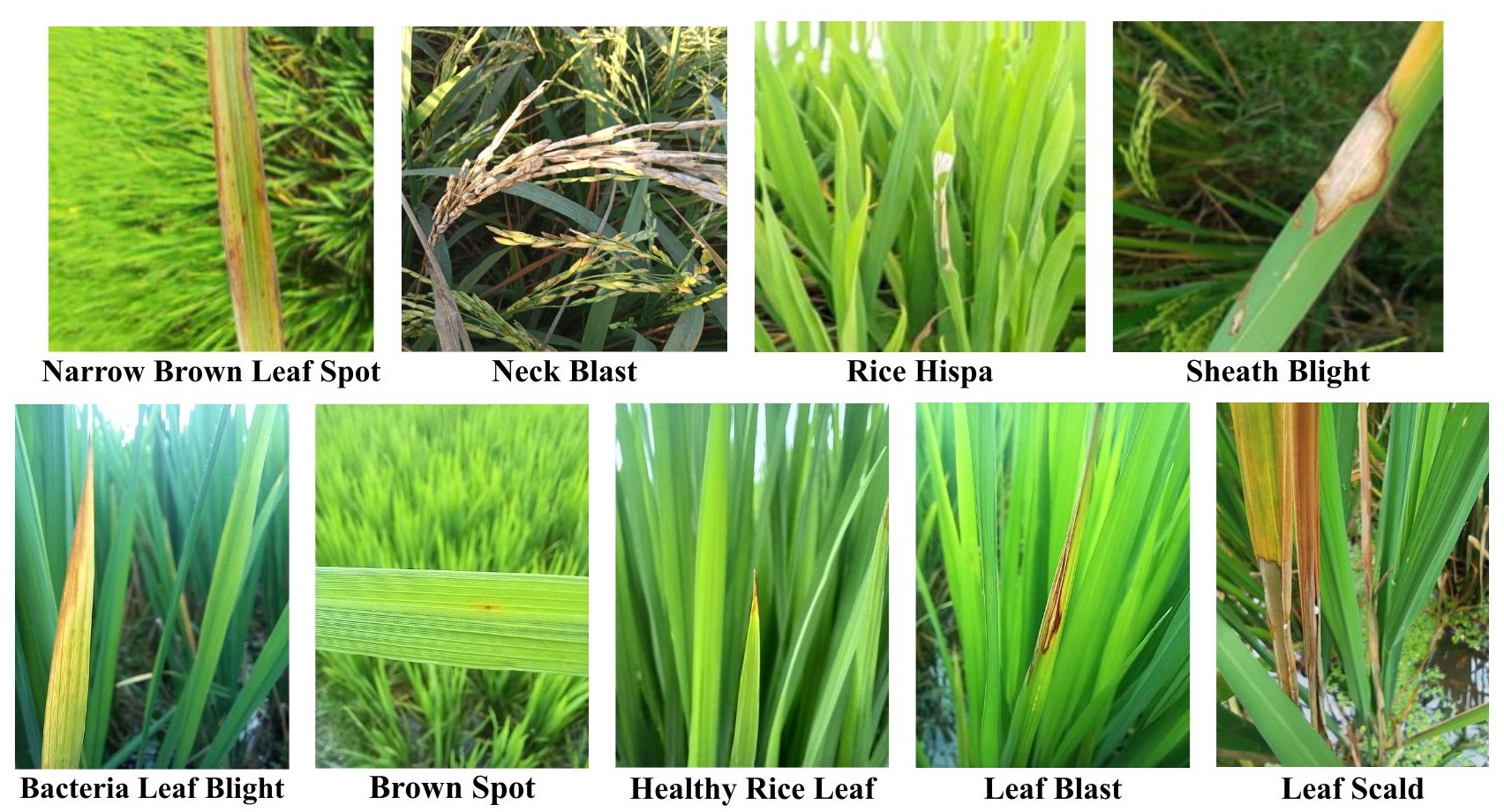}
    \caption{Sample images from the rice leaf disease dataset showing various leaf diseases.}
    \label{fig:rice_leaf_dataset}
\end{figure}

This dataset (\href{https://www.kaggle.com/datasets/loki4514/rice-leaf-diseases-detection}{Kaggle dataset}) is an enhanced collection of 11{,}790 labeled images of rice leaves, encompassing both healthy and diseased conditions. The dataset is provided in an augmented form by the authors, incorporating various transformations such as rotation, scaling, horizontal and vertical flipping, and color manipulation to increase sample diversity and robustness. This diversity supports effective model training for distinguishing visual patterns across different leaf states. It includes nine categories: \textit{Leaf Blast} (1{,}748), \textit{Sheath Blight} (1{,}629), \textit{Brown Spot} (1{,}546), \textit{Leaf Scald} (1{,}332), \textit{Rice Hispa} (1{,}299), \textit{Bacterial Leaf Blight} (1{,}197), \textit{Healthy Rice Leaf} (1{,}085), \textit{Neck Blast} (1{,}000), and \textit{Narrow Brown Leaf Spot} (954). The dataset is split into 80\%, 10\%, and 10\% for training, validation, and testing, respectively. Sample images illustrating these conditions are presented in Figure~\ref{fig:rice_leaf_dataset}.

\subsubsection{Potato leaf disease dataset}

\begin{figure}[H]
    \centering
    \includegraphics[width=0.7\textwidth]{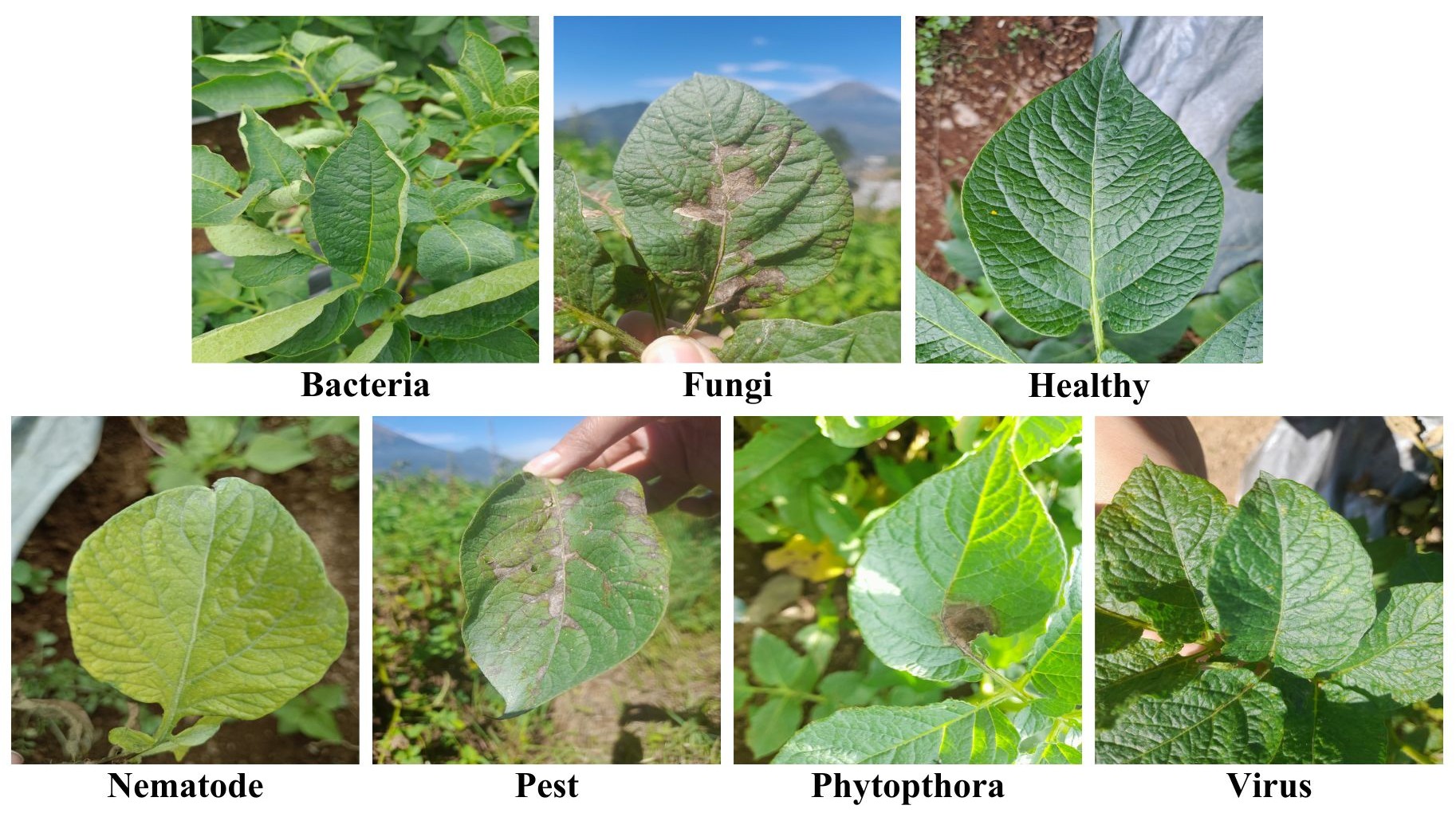}
    \caption{Sample images from the potato leaf disease dataset showing different leaf diseases.}
    \label{fig:potato_leaf_dataset}
\end{figure}

This dataset is a comprehensive collection of 3,076 labeled images of potato leaves, captured from farms in Central Java, Indonesia~\citep{Shabrina2024}. The images belong to seven distinct classes: \textit{Healthy} (201 images), \textit{Bacteria} (569 images), \textit{Fungi} (748 images), \textit{Nematode} (68 images), \textit{Pest} (611 images), \textit{Phytophthora} (347 images), and \textit{Virus} (532 images). All images are provided in \texttt{JPEG} format with a high resolution of $1500 \times 1500$ pixels. The dataset is partitioned into 81\%, 9\%, and 10\% for training, validation, and testing, respectively, following the original study~\citep{Shabrina2024}. Sample images illustrating these conditions are presented in Figure~\ref{fig:potato_leaf_dataset}.

\subsubsection{Coffee leaf disease dataset}

\begin{figure}[H]
    \centering
    \includegraphics[width=0.7\textwidth]{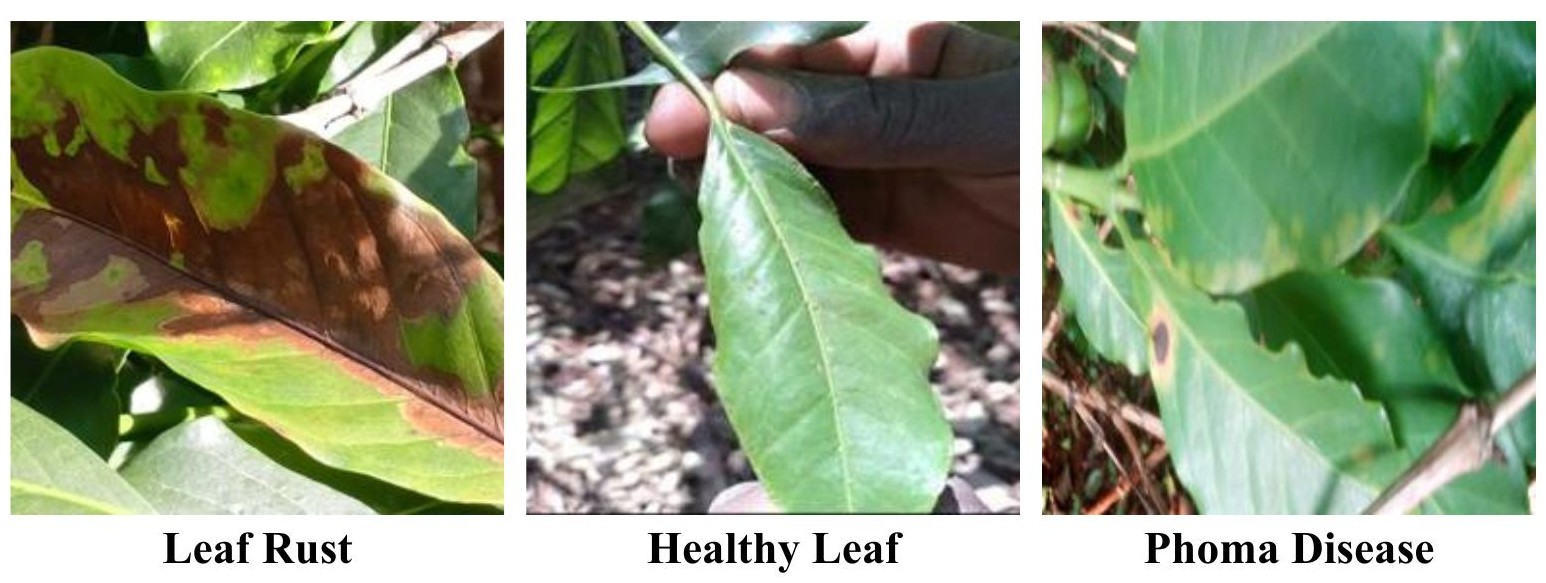}
    \caption{Sample images from the coffee leaf disease dataset illustrating different leaf diseases.}
    \label{fig:coffee_leaf_dataset}
\end{figure}

This dataset is a structured collection of 3,219 labeled images of coffee leaves, captured from farms in Uganda~\citep{Chelangat2025}. The images belong to three distinct classes: \textit{Healthy} (1,078 images), \textit{Leaf Rust} (1,031 images), and \textit{Phoma Disease} (1,110 images). All images are provided in \texttt{JPEG} format with a consistent resolution of $256 \times 256$ pixels. The dataset is divided into 64\%, 16\%, and 20\% for training, validation, and testing, respectively. Sample images illustrating these conditions are presented in Figure~\ref{fig:coffee_leaf_dataset}.

\subsubsection{Corn leaf disease dataset}

\begin{figure}[H]
    \centering
    \includegraphics[width=1.0\textwidth]{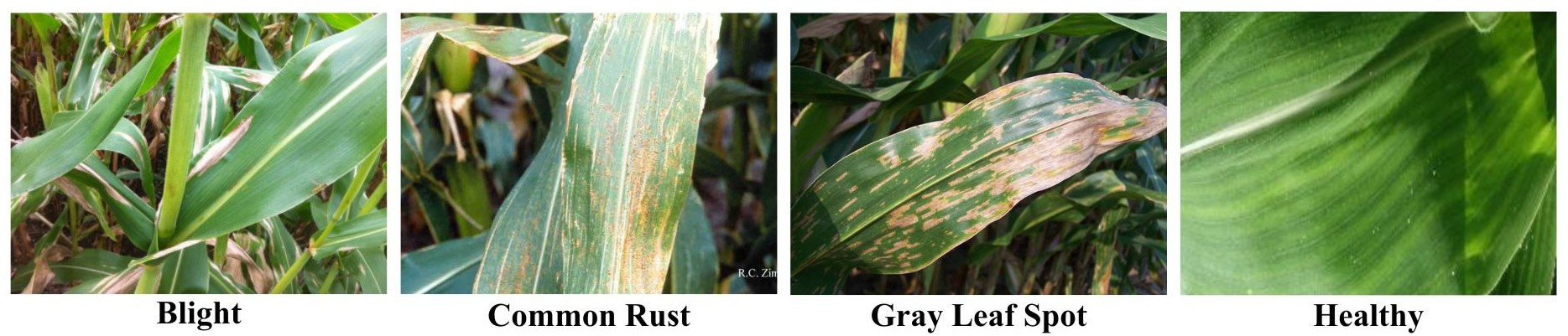}
    \caption{Sample images from the corn leaf disease dataset capturing various leaf disease patterns.}
    \label{fig:corn_leaf_dataset}
\end{figure}

This dataset is a collection of 4,188 images sourced from the PlantVillage dataset, focusing on common corn leaf diseases~\citep{ArunPandian2019}. The images are categorized into four distinct classes: \textit{Common Rust} (1,306 images), \textit{Blight} (1,146 images), \textit{Gray Leaf Spot} (574 images), and \textit{Healthy} (1,162 images). The dataset is split into 64\%, 16\%, and 20\% for training, validation, and testing, respectively. Sample images illustrating these conditions are presented in Figure~\ref{fig:corn_leaf_dataset}.

\subsection{Experimental setting}

All experiments are conducted on the Kaggle platform using two NVIDIA T4 GPUs.
To ensure fair comparison and reproducibility, a unified experimental environment and consistent training protocol are adopted across all datasets.
Unless otherwise specified, all models are trained for 100 epochs using a batch size of 64 and optimized with the Adam optimizer at a learning rate of $10^{-4}$. Early stopping is applied with a patience of 20 epochs to mitigate overfitting. For the Corn leaf disease and Potato leaf disease datasets, smaller batch sizes of 8 and 4 are adopted, respectively, to better accommodate dataset characteristics and improve training stability under class imbalance. In addition, AdamW ($\text{learning rate}=10^{-4}$, weight decay = $10^{-2}$) is employed for the Potato leaf disease dataset to enhance regularization, while all other training settings remain unchanged.
The detailed configurations for data preprocessing and loss hyperparameters for each dataset are summarized in Table~\ref{tab:exp_configs}.

{\footnotesize
\begin{longtable}{|p{1.5cm}|p{5.5cm}|p{3.9cm}|}
\caption{Detailed experimental configurations for each dataset.}
\label{tab:exp_configs}\\
\hline
\textbf{Dataset} &
\textbf{Preprocessing \& augmentation} &
\textbf{Loss hyperparameters} $(\lambda_1, \lambda_2, \lambda_3, \lambda_4, \tau, \tau')$ \\
\hline
\endfirsthead

\multicolumn{3}{c}%
{{\tablename\ \thetable{} -- continued from previous page}} \\
\hline
\textbf{Dataset} &
\textbf{Preprocessing \& augmentation} &
\textbf{Loss hyperparameters} $(\lambda_1, \lambda_2, \lambda_3, \lambda_4, \tau, \tau')$ \\
\hline
\endhead

\hline
\endlastfoot

\textbf{Rice variety} &
\begin{tabular}[t]{@{}l@{}}
Resize $224 \times 224$\\
Random horizontal flip\\
Random vertical flip\\
Random rotation ($\pm10^\circ$)
\end{tabular} &
$[0.3, 0.7, 0.7, 0.7, 4, 4]$ \\
\hline

\textbf{Rice leaf disease} &
\begin{tabular}[t]{@{}l@{}}
Resize $224 \times 224$\\
Color jitter\\
Random affine\\
Random rotation ($\pm20^\circ$)
\end{tabular} &
$[0.3, 0.7, 0.7, 0.7, 4, 4]$ \\
\hline

\textbf{Potato leaf disease} &
\begin{tabular}[t]{@{}l@{}}
Resize $224 \times 224$\\
Color jitter\\
Random affine\\
Random rotation ($\pm30^\circ$)
\end{tabular} &
$[0.2, 1.0, 0.8, 0.8, 4, 4]$ \\
\hline

\textbf{Coffee leaf disease} &
\begin{tabular}[t]{@{}l@{}}
Resize $224 \times 224$\\
Color jitter\\
Random affine\\
Random rotation ($\pm20^\circ$)
\end{tabular} &
$[0.3, 0.7, 0.7, 0.7, 4, 4]$ \\
\hline

\textbf{Corn leaf disease} &
\begin{tabular}[t]{@{}l@{}}
Resize $224 \times 224$\\
Random horizontal flip\\
Random vertical flip\\
Random rotation ($\pm30^\circ$)
\end{tabular} &
$[0.3, 0.7, 0.7, 0.7, 4, 4]$ \\

\end{longtable}

\textit{Note:}
$(\lambda_1, \lambda_2, \lambda_3, \lambda_4)$ denote the loss weighting coefficients, while
$\tau$ and $\tau'$ represent the temperature parameters used in the distillation process.
}

\section{Results}\label{sec:results}

This section presents a comprehensive evaluation of the proposed framework across multiple agricultural datasets. The experiments are organized to progressively analyze the effectiveness of the method from different perspectives. We begin by benchmarking several representative deep learning architectures to establish a performance reference for the rice variety recognition task. Based on these observations, subsequent experiments focus on evaluating the proposed knowledge distillation framework, including efficiency–accuracy trade-offs and the impact of different design choices. Ablation studies are conducted to examine the contribution of each distillation component, followed by qualitative analysis using visualization techniques to interpret model behavior. Finally, cross-dataset evaluations on plant leaf disease benchmarks are performed to assess the robustness and generalization capability of the proposed approach.

\subsection{Performance of pre-trained models for rice seed purity recognition}\label{sec:benchmark_results}

To establish a reliable performance baseline for the rice variety identification task, we conducted a comparative evaluation using a diverse set of popular deep learning models. The selected architectures cover a broad spectrum of design philosophies, including conventional convolutional networks (VGG, ResNet), densely connected models (DenseNet), efficiency-oriented architectures (MobileNet, EfficientNet), modernized convolutional designs (ConvNeXt), and Transformer-based approaches (ViT, Swin-Tiny).

The results in Table~\ref{tab:baseline_result} indicate that the dataset exhibits consistent discriminative patterns across different model families, with all evaluated architectures achieving strong recognition performance. More importantly, the observed performance variations among these models provide valuable insights into how architectural complexity and inductive biases influence recognition accuracy on this task. These observations serve as a principled reference for the subsequent design and evaluation of the proposed knowledge distillation framework.

Among the evaluated architectures, DenseNet121 achieved the highest mean accuracy of 98.55\%, closely followed by EfficientNetB0 at 98.52\%. The strong performance of DenseNet121 can be attributed to its dense connectivity, which promotes feature reuse and facilitates gradient flow, enabling the capture of fine-grained textural details that are critical for distinguishing visually similar rice varieties. EfficientNetB0 further demonstrates the effectiveness of compound scaling, which jointly balances network depth, width, and input resolution to optimize feature extraction. ResNet18 also achieved a competitive accuracy of 98.50\%, highlighting the role of residual connections in supporting stable optimization and robust representation learning. Collectively, these results establish a strong performance benchmark and indicate that modern CNN architectures are well suited for fine-grained rice variety identification.

Further analysis across different CNN families reinforces this observation. Modernized architectures such as ConvNeXt achieved an accuracy of 98.27\%, while lightweight models such as MobileNetV2 reached 98.02\%, demonstrating that high recognition performance can be maintained even with compact network designs. In contrast, the VGG16 model achieved an accuracy of 97.87\%, and its relatively lower performance compared to more recent architectures reflects the evolution of CNN design toward more efficient and expressive representations.

\begin{landscape}

\begin{table}[H]
\centering
\caption{Performance of pre-trained models for rice seed purity identification (\%).}
\label{tab:baseline_result}
\renewcommand{\arraystretch}{1.3}
\setlength{\tabcolsep}{5pt}
\makebox[\textwidth][c]{
    \resizebox{1.4\textwidth}{!}{ 
        \begin{tabular}{|c|c|c|c|c|c|c|c|c|}
        \hline
        \textbf{\shortstack{Rice seed \\ variety}} 
        & \textbf{MobileNetV2} 
        & \textbf{EfficientNetB0}
        & \textbf{DenseNet121}
        & \textbf{ResNet18} 
        & \textbf{Swin-Tiny} 
        & \textbf{ViT} 
        & \textbf{ConvNeXt} 
        & \textbf{VGG16} 
        \\
        \hline
        BC-15        & 98.40 & 98.80 & 98.80 & 98.94 & 98.54 & 98.80 & 98.01 & 97.74 \\ \hline
        Huong Thom-1 & 98.15 & 99.07 & 99.31 & 98.96 & 98.96 & 97.92 & 98.61 & 98.50 \\ \hline
        Nep-87       & 99.30 & 99.48 & 99.30 & 99.65 & 98.43 & 97.91 & 98.78 & 98.78 \\ \hline
        Q-5          & 98.35 & 97.34 & 98.23 & 98.73 & 97.72 & 97.21 & 97.85 & 97.97 \\ \hline
        TBR-36       & 97.85 & 99.57 & 98.28 & 99.57 & 98.50 & 99.14 & 98.93 & 97.85 \\ \hline
        TBR-45       & 96.58 & 98.29 & 97.44 & 97.01 & 98.50 & 96.79 & 96.79 & 96.37 \\ \hline
        TH-35        & 97.83 & 97.59 & 99.28 & 97.35 & 98.07 & 97.35 & 98.55 & 98.31 \\ \hline
        Thien Uu-8   & 96.91 & 97.62 & 97.15 & 97.62 & 97.39 & 96.91 & 97.39 & 96.20 \\ \hline
        Xi-23        & 98.80 & 98.91 & 99.12 & 98.69 & 99.34 & 98.47 & 99.56 & 99.12 \\ \hline
        \textbf{\shortstack{Mean \\ Accuracy}} 
        & \textbf{98.02} & \textbf{98.52} & \textbf{98.55} & \textbf{98.50} & \textbf{98.38} & \textbf{97.83} & \textbf{98.27} & \textbf{97.87} \\  \hline
        \end{tabular}
    }
}
\end{table}
                              
Transformer-based models exhibited a distinct performance trend compared to CNN architectures (Table~\ref{tab:baseline_result}). Specifically, Swin-Tiny achieved an accuracy of 98.38\%, while ViT reached 97.83\%. Although their performance is competitive, these results suggest that CNN-based models remain particularly effective for this fine-grained classification task. A plausible explanation lies in the inductive bias of CNNs toward spatial locality, which facilitates the learning of fine textures and local visual patterns that are critical for distinguishing visually similar rice varieties. In contrast, Transformer-based models rely primarily on global self-attention mechanisms and typically require larger-scale datasets to capture such localized details effectively.

\end{landscape}

These observations provide empirical motivation for the proposed knowledge distillation framework. The consistently strong performance of ResNet18, together with its favorable balance between architectural simplicity and representational capacity, supports its selection as the teacher model. Furthermore, the complementary strengths demonstrated by DenseNet121 in feature representation and by MobileNetV2 in computational efficiency motivate the incorporation of dense connectivity and inverted residual structures into the student model design.

\subsection{Performance of the proposed distillation framework on the rice seed varieties dataset}

This section evaluates the proposed hybrid knowledge distillation framework on the rice variety identification task. The performance of both the teacher and student models, including their main and auxiliary branches, is analyzed to assess the effectiveness of knowledge transfer to the lightweight student network. The results summarized in Table~\ref{tab:KD_result} provide the basis for evaluating the accuracy and generalization capability of the proposed method across different rice varieties.

\begin{table}[H]
\centering
\caption{Performance of the proposed distillation framework on the rice seed varieties dataset (\%).}
\label{tab:KD_result}
\resizebox{\textwidth}{!}{
\begin{tabular}{|c|c|c|c|c|}
\hline
\textbf{Rice seed variety} & \textbf{Student auxiliary} & \textbf{Student main} & \textbf{Teacher auxiliary} & \textbf{Teacher main} \\
\hline
BC-15           & 98.67                     & 99.07                     & 99.20                    & 99.07                     \\ \hline
Huong Thom-1    & 97.69                     & 98.15                     & 99.19                    & 99.19                     \\ \hline
Nep-87          & 99.65                     & 99.65                     & 99.83                    & 99.65                     \\ \hline
Q-5             & 98.35                     & 98.48                     & 98.35                    & 98.23                     \\ \hline
TBR-36          & 99.36                     & 99.36                     & 99.57                    & 99.79                     \\ \hline
TBR-45          & 97.22                     & 97.22                     & 95.73                    & 96.79                     \\ \hline
TH-35           & 99.28                     & 98.80                     & 98.31                    & 99.04                     \\ \hline
Thien Uu-8      & 97.86                     & 97.86                     & 97.15                    & 96.67                     \\ \hline
Xi-23           & 98.25                     & 98.47                     & 99.02                     & 99.45                     \\ \hline
\textbf{Mean Accuracy} & \textbf{98.48}    & \textbf{98.56}            & \textbf{98.48}            & \textbf{98.65}            \\
\hline
\end{tabular}
}
\end{table}

The results in Table~\ref{tab:KD_result} indicate that the proposed hybrid knowledge distillation framework achieves strong and consistent performance on the rice variety identification task. The student model attains a mean accuracy of 98.56\% on the main branch, with only a marginal difference of 0.09\% compared to the teacher main branch (98.65\%), while the student auxiliary branch reaches 98.48\%, matching the performance of the teacher auxiliary branch. Moreover, the student model outperforms the teacher on several visually challenging rice varieties, such as Thien Uu-8 and TBR-45, which exhibit highly similar morphological characteristics and subtle inter-class differences. For the TH-35 variety, the student auxiliary branch achieves the highest accuracy of 99.28\%, further highlighting the effectiveness of intermediate supervision in facilitating the learning of fine-grained and discriminative representations. 

This analysis confirms that the distillation framework enables effective knowledge transfer and robust feature learning in a lightweight student network, while reducing the influence of teacher-specific biases. The student achieves performance comparable to the teacher while demonstrating improved generalization on rice varieties with subtle morphological differences. The combination of architectural complementarity and multi-level distillation supports effective high-level class discrimination and preservation of low-level, texture-rich details, making the framework well-suited for deployment in resource-constrained environments.

\subsection{Analysis of knowledge distillation effects}

To gain deeper insight into the effects of the proposed hybrid knowledge distillation framework, experiments are conducted on the rice variety identification task. The analysis considers classification performance, model complexity, and visual interpretability, including accuracy and computational efficiency evaluation, an ablation study on the distillation loss components, and a qualitative analysis of attention behavior using Grad-CAM.

\subsubsection{Evaluation of model accuracy and computational efficiency}

\begin{figure}[H]
    \hspace{-2.7cm}
    \includegraphics[width=1.35\textwidth]{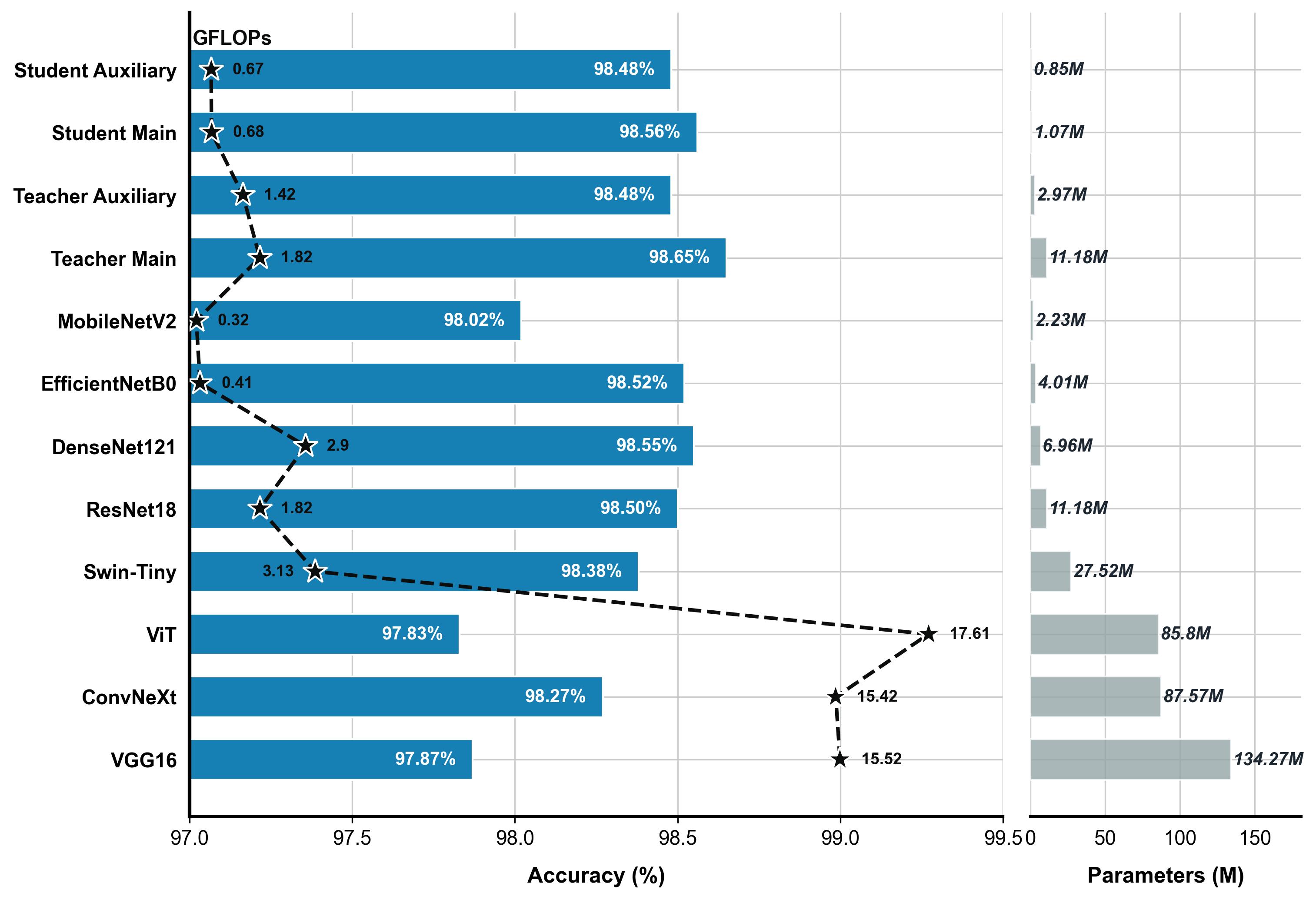}
    \caption{Comprehensive comparison of accuracy (\%), GFLOPs, and parameter count between the proposed distillation framework and the popular pretrained models on the rice-variety dataset}
    \label{fig:comparison_acc_params_flops}
\end{figure}

The performance of the proposed knowledge distillation method is assessed by comparing the student model with its teacher model and a set of pretrained baseline models evaluated in Section~\ref{sec:benchmark_results} on the rice variety identification dataset. The comparison focuses on three key aspects: classification accuracy, computational cost measured in GFLOPs, and model size in terms of the number of parameters. A visual summary of the comparative results is presented in Figure~\ref{fig:comparison_acc_params_flops}.

The accuracy analysis indicates that the main distilled student model achieves highly competitive performance despite its compact architecture. Specifically, the main student branch attains an accuracy of 98.56\%, which is only marginally lower than that of the main teacher model at 98.65\%, corresponding to a negligible gap of 0.09\%. This result demonstrates that the proposed distillation strategy successfully transfers nearly all of the teacher's discriminative capability to the student model. When compared with the pretrained baseline models, the main student branch outperforms lightweight architectures such as MobileNetV2 (98.02\%) and achieves comparable or slightly better performance than EfficientNetB0 (98.52\%) and ResNet18 (98.50\%). Notably, its accuracy is comparable to that of DenseNet121 (98.55\%), positioning the distilled student model among the best-performing approaches within the evaluated pretrained models. These results indicate that the proposed knowledge distillation framework effectively mitigates the trade-off between accuracy and computational efficiency, enabling a compact model to deliver highly competitive predictive performance.

Notably, the teacher model equipped with the auxiliary branch improves the classification accuracy from 98.50\%, obtained with the standard ResNet18, to 98.65\%. This gain of 0.15\% highlights the effectiveness of the auxiliary branch in providing intermediate supervision during training. Specifically, the auxiliary branch introduces an additional loss term computed from intermediate feature representations, which supplies a stronger and more direct supervisory signal to deeper layers of the network. Beyond facilitating feature learning at multiple depths, this auxiliary supervision enhances gradient propagation, encourages the extraction of richer and more discriminative representations, and enables the model to capture both low-level texture patterns and high-level semantic cues. Collectively, these factors contribute to improved robustness in classification, particularly for fine-grained distinctions among rice varieties.

Furthermore, the proposed framework exhibits a clear advantage in computational efficiency. The main student branch requires only 0.68 GFLOPs, indicating a substantially lower computational cost than the evaluated baselines. Specifically, it is approximately 2.7 times more computationally efficient than the teacher branch (1.82 GFLOPs), 4.2 times more efficient than DenseNet121 (2.9 GFLOPs), and more than 25 times more efficient than transformer-based architectures such as ViT.
In terms of model size, the main student network is highly compact, containing only 1.07 million parameters. This corresponds to an approximately 10.4 times reduction compared with the main teacher network (11.18M parameters) and results in a smaller footprint than other lightweight models, including MobileNetV2 (2.23M parameters) and EfficientNetB0 (4.01M parameters). Moreover, when compared with large-scale convolutional models such as VGG16 (134.27M parameters), the proposed student model is over 100 times smaller.
These results highlight the suitability of the proposed student architecture for deployment in resource-constrained environments, where both computational efficiency and model compactness are critical.

Considering accuracy, computational cost, and model size jointly, the proposed knowledge distillation framework yields a student model that achieves a favorable balance between predictive performance and efficiency. The distilled student preserves nearly the full classification capability of the teacher while substantially reducing both computational complexity and memory requirements. This balance highlights the practical potential of the proposed approach, particularly for smart agriculture applications where on-device or edge deployment is essential.

\subsubsection{Impact of distillation loss terms (Ablation study)}

To demonstrate the robustness of the proposed distillation framework under challenging fine-grained conditions, ablation experiments were conducted on the TBR-45 rice variety, which represents the most difficult class in the dataset due to its high visual similarity to other varieties. The results, summarized in Table~\ref{tab:ablation_study}, quantify the contribution of each distillation loss term and justify the design choices of the final training objective. 

\begin{table}[H]
\footnotesize
\centering
\caption{Ablation study analyzing the impact of each loss component on the classification accuracy of the TBR-45 rice variety.}
\label{tab:ablation_study}
\setlength{\tabcolsep}{3pt}
\begin{tabular}{|l|c|c|c|c|c|c|c|}
\hline
\multirow{2}{*}{\textbf{Model variant}} & \multirow{2}{*}{$\mathcal{L}_{\text{Hard}}$} & \multirow{2}{*}{$\mathcal{L}_{\text{RD}}$} & \multirow{2}{*}{$\mathcal{L}_{\text{FD}}$} & \multirow{2}{*}{$\mathcal{L}_{\text{SD}}$} & \multicolumn{2}{c|}{\textbf{Accuracy (\%)}} & \multirow{2}{*}{\textbf{Gain} (\%)} \\ \cline{6-7}
 & & & & & \textbf{Auxiliary} & \textbf{Main} & \\ \hline
Baseline (Standard KD) & \checkmark & \checkmark & - & - & 94.23 & 94.23 & - \\ \hline
Baseline + $\mathcal{L}_{\text{FD}}$ (MSE) & \checkmark & \checkmark & \checkmark & - & 94.23 & 95.09 & +0.86 \\ \hline
Baseline + $\mathcal{L}_{\text{FD}}$ (Euclidean) & \checkmark & \checkmark & \checkmark & - & 94.87 & 95.94 & +1.71 \\ \hline
\textbf{Proposed method} & \textbf{\checkmark} & \textbf{\checkmark} & \textbf{\checkmark} & \textbf{\checkmark} & \textbf{97.22} & \textbf{97.22} & \textbf{+2.99} \\ \hline
\end{tabular}
\end{table}

The analysis begins by establishing a performance baseline using only hard-label supervision ($\mathcal{L}_{\text{Hard}}$) and response-based distillation ($\mathcal{L}_{\text{RD}}$), which achieves an accuracy of 94.23\%. Hard-label supervision guides the model to learn directly from ground-truth annotations, while response-based distillation enables the student to mimic the teacher's soft outputs, thereby capturing inter-class relationships encoded by the teacher.

In the next phase, we incorporated the feature-based distillation loss ($\mathcal{L}_{\text{FD}}$) to encourage the student model to mimic not only the teacher's final predictions but also its internal feature representations. When $\mathcal{L}_{\text{FD}}$ was implemented using Mean Squared Error (MSE), the main branch accuracy improved to 95.09\% (+0.86\%). Notably, replacing MSE with the Euclidean distance yielded a more substantial performance gain, achieving 95.94\% (+1.71\%). The superiority of the Euclidean distance indicates that directly penalizing the magnitude of differences between feature vectors provides a more stable and effective supervisory signal for representation alignment in this task. Furthermore, $\mathcal{L}_{\text{FD}}$ was applied to the feature vectors extracted immediately after the Global Average Pooling (GAP) layer, which implicitly acts as a form of channel-wise attention guidance. This design encourages the student model to prioritize the most informative feature channels emphasized by the teacher, rather than rigidly matching spatial feature maps.

Finally, building upon the best-performing configuration, we introduced the self-distillation loss ($\mathcal{L}_{\text{SD}}$) to complete the proposed hybrid framework. This mechanism enables intra-network knowledge transfer, with the shallower auxiliary branch serving as an ``internal mentor'' to regularize the deeper main branch and anchor its representations to the foundational visual features learned in earlier layers. Integrating this component increased the model's accuracy to 97.22\%, representing a total improvement of +2.99\% over the initial baseline and an additional gain of +1.28\% over the previous configuration (Baseline + $\mathcal{L}_{\text{FD}}$). This result highlights the critical and unique contribution of self-distillation in complementing feature- and response-based distillation, ultimately enabling the proposed hybrid loss to achieve optimal performance.

\subsubsection{Visual interpretation of model attention using Grad-CAM}

\begin{figure}[!ht]
    \centering

    \begin{subfigure}[b]{0.35\textwidth}
        \centering
        \includegraphics[width=\linewidth]{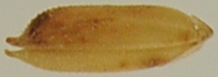} 
        \caption{Original input image}
        \label{fig:input}
    \end{subfigure}

    \vspace{0.8em}
    
    \begin{subfigure}[b]{0.245\textwidth}
        \centering
        \includegraphics[width=\linewidth]{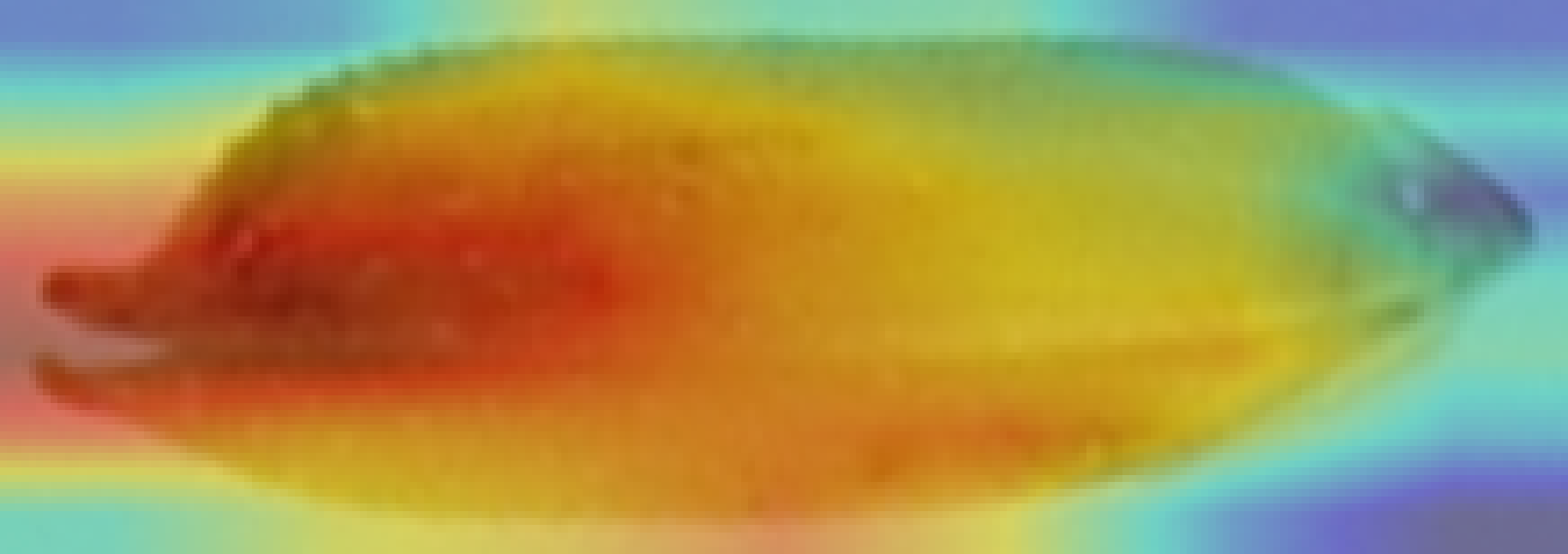} 
        \caption{Student main}
        \label{fig:student_main_grad_cam}
    \end{subfigure}\hfill
    \begin{subfigure}[b]{0.245\textwidth}
        \centering
        \includegraphics[width=\linewidth]{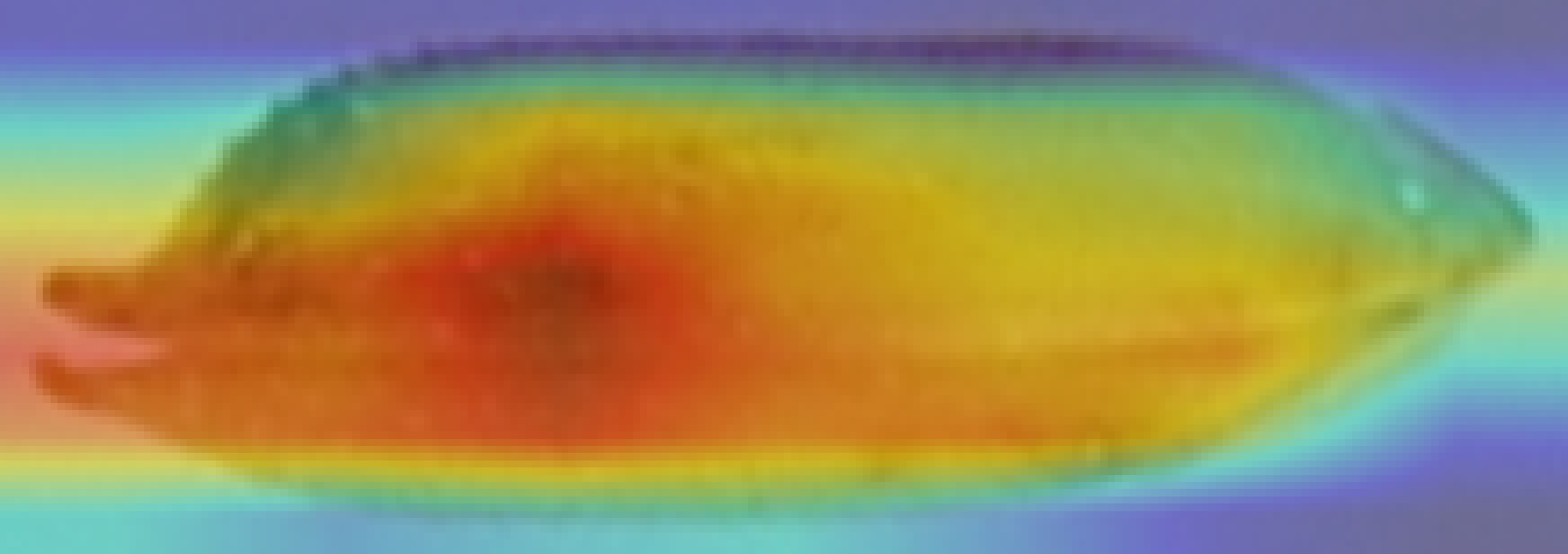}
        \caption{Student auxiliary}
        \label{fig:student_aux_grad_cam}
    \end{subfigure}\hfill
    \begin{subfigure}[b]{0.245\textwidth}
        \centering
        \includegraphics[width=\linewidth]{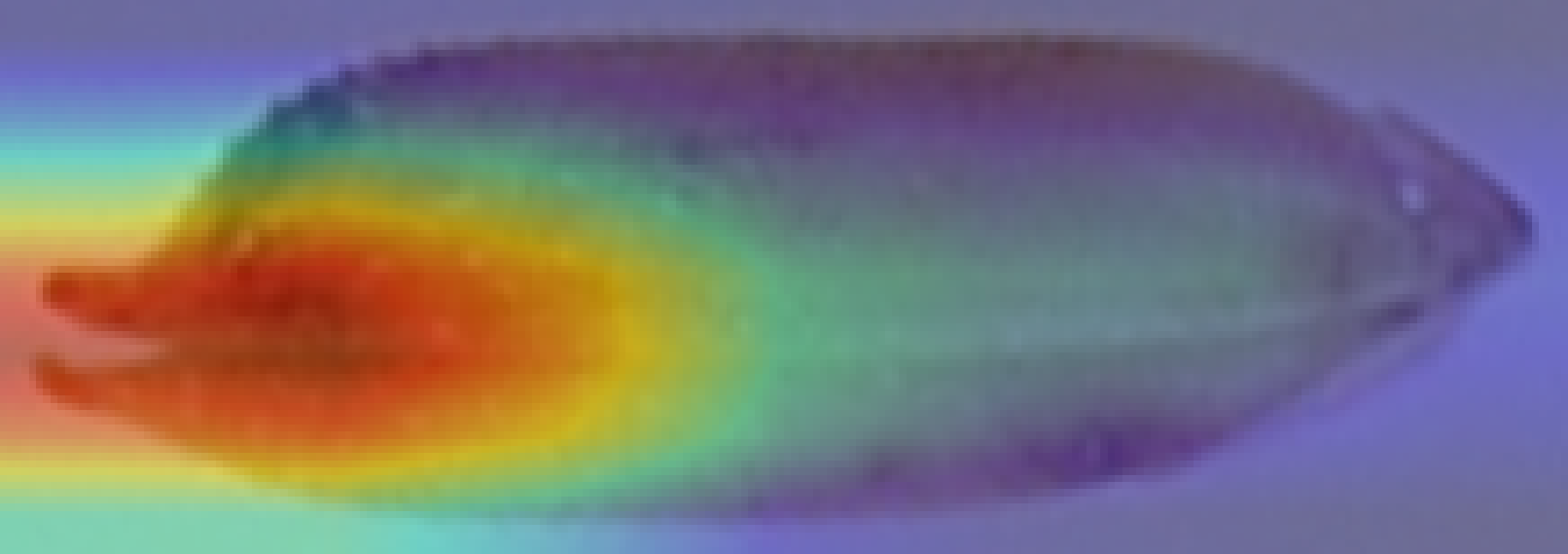}
        \caption{Teacher main}
        \label{fig:teacher_main_grad_cam}
    \end{subfigure}\hfill
    \begin{subfigure}[b]{0.245\textwidth}
        \centering
        \includegraphics[width=\linewidth]{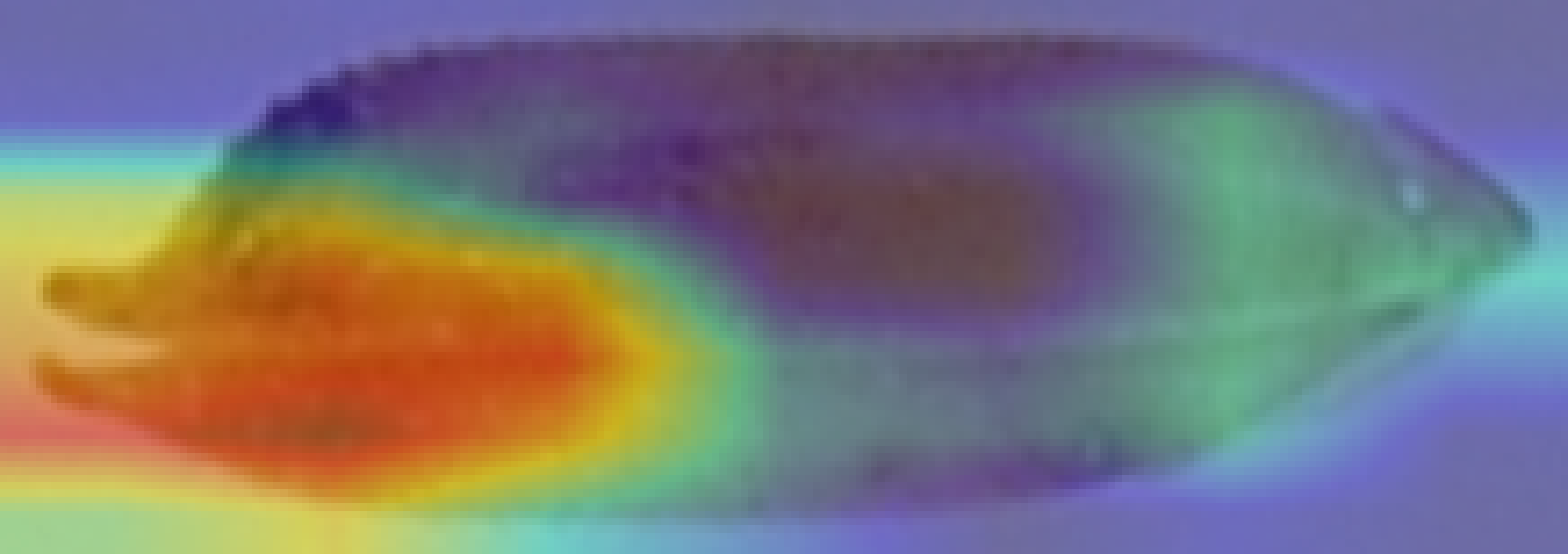}
        \caption{Teacher auxiliary}
        \label{fig:teacher_aux_grad_cam}
    \end{subfigure}
    
    \vspace{0.8em}

    \begin{subfigure}[b]{0.245\textwidth}
        \centering
        \includegraphics[width=\linewidth]{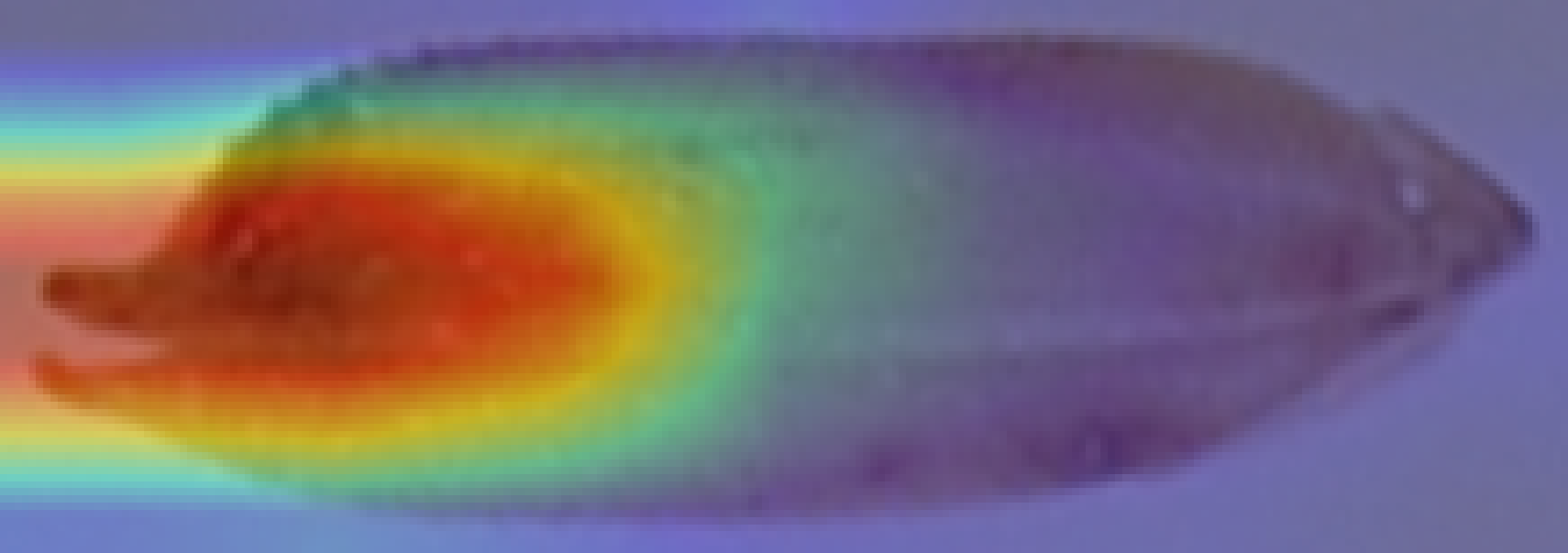}
        \caption{MobileNetV2}
        \label{fig:mobileNet_grad_cam}
    \end{subfigure}\hfill
    \begin{subfigure}[b]{0.245\textwidth}
        \centering
        \includegraphics[width=\linewidth]{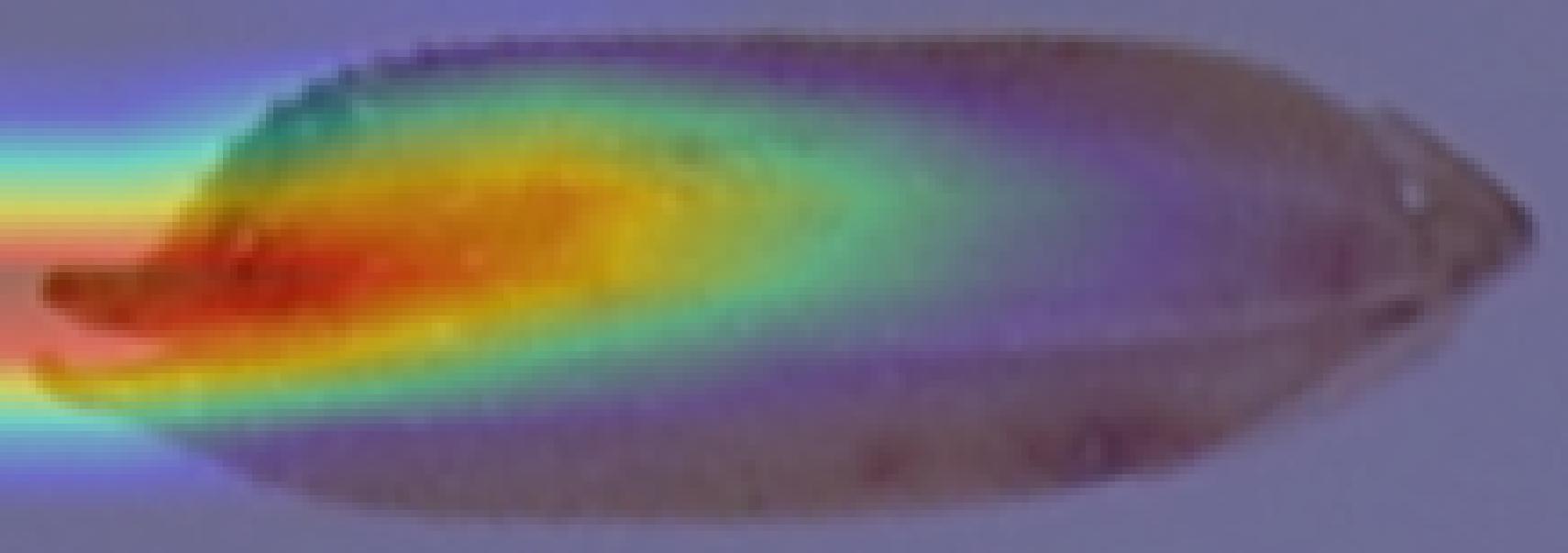}
        \caption{EfficientNetB0}
        \label{fig:efficientNet_grad_cam}
    \end{subfigure}\hfill
    \begin{subfigure}[b]{0.245\textwidth}
        \centering
        \includegraphics[width=\linewidth]{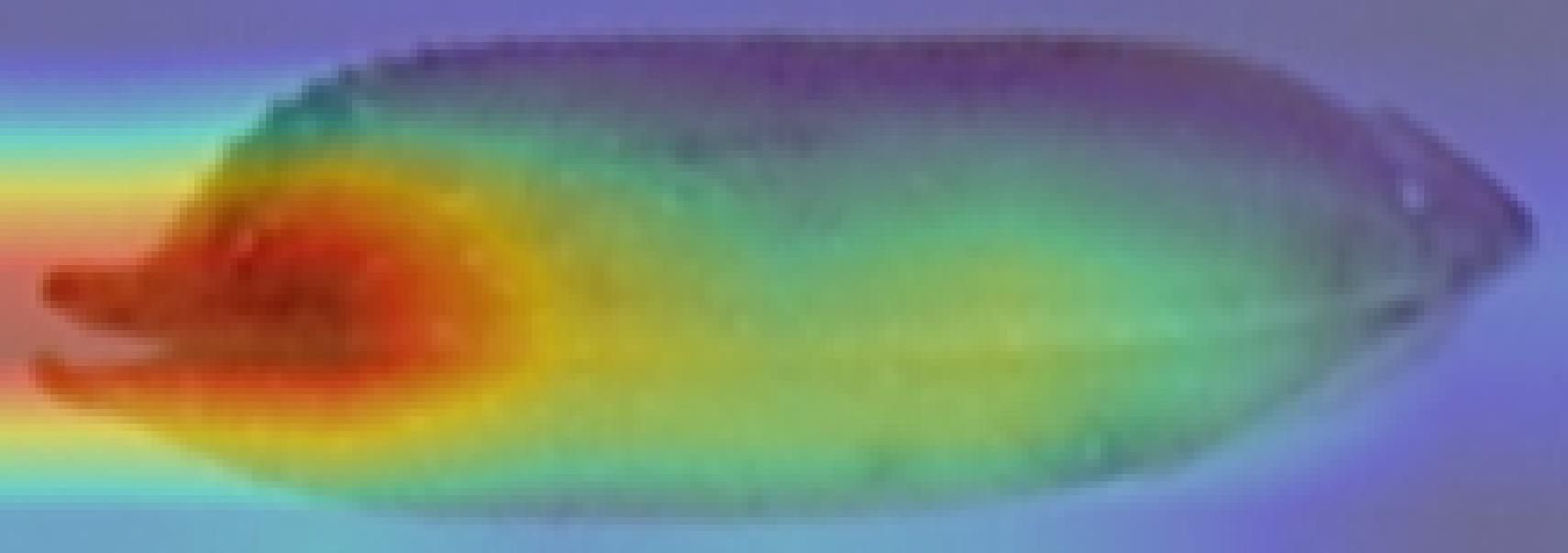}
        \caption{DenseNet121}
        \label{fig:densenet_grad_cam}
    \end{subfigure}\hfill
    \begin{subfigure}[b]{0.245\textwidth}
        \centering
        \includegraphics[width=\linewidth]{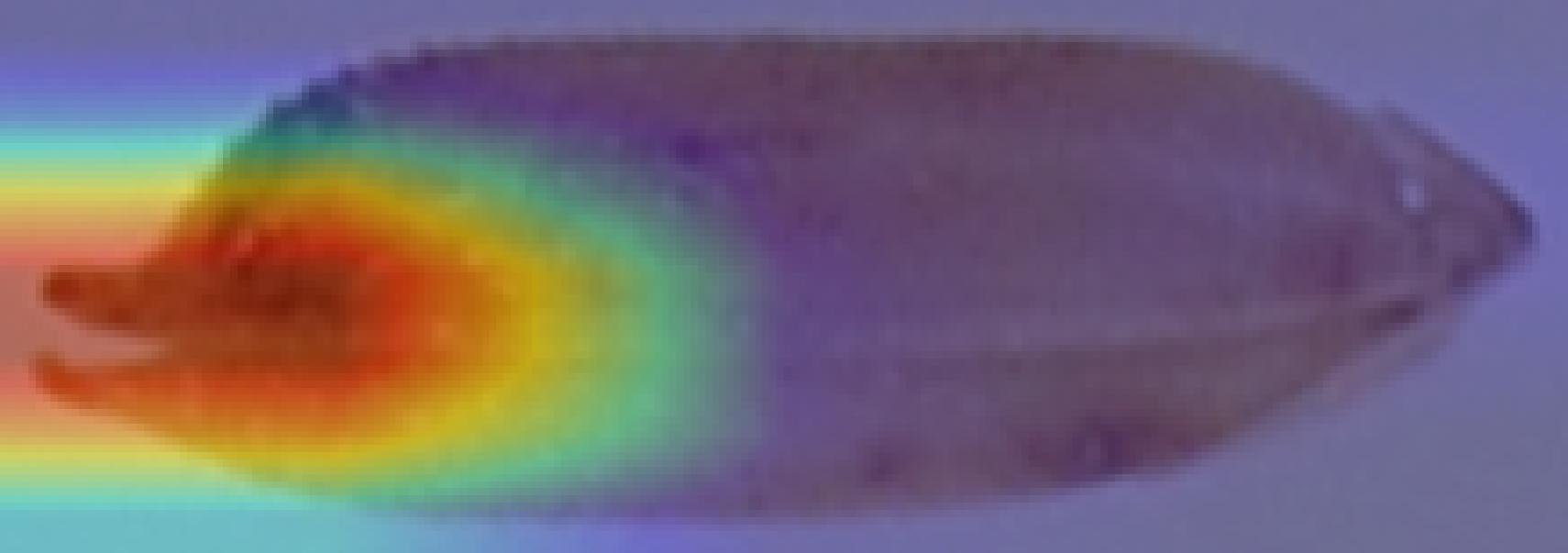}
        \caption{ResNet18}
        \label{fig:resnet_grad_cam}
    \end{subfigure}

    \vspace{0.8em} 

    \begin{subfigure}[b]{0.245\textwidth}
        \centering
        \includegraphics[width=\linewidth]{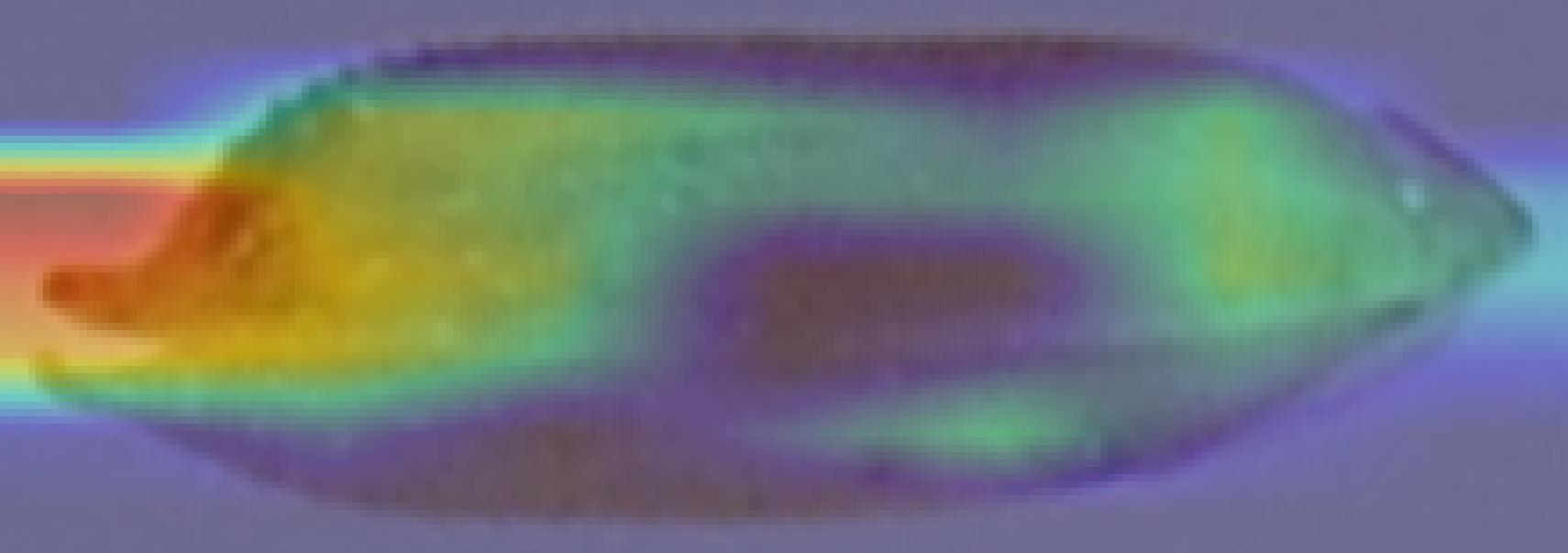}
        \caption{Swin-Tiny}
        \label{fig:swin_grad_cam}
    \end{subfigure}\hfill
    \begin{subfigure}[b]{0.245\textwidth}
        \centering
        \includegraphics[width=\linewidth]{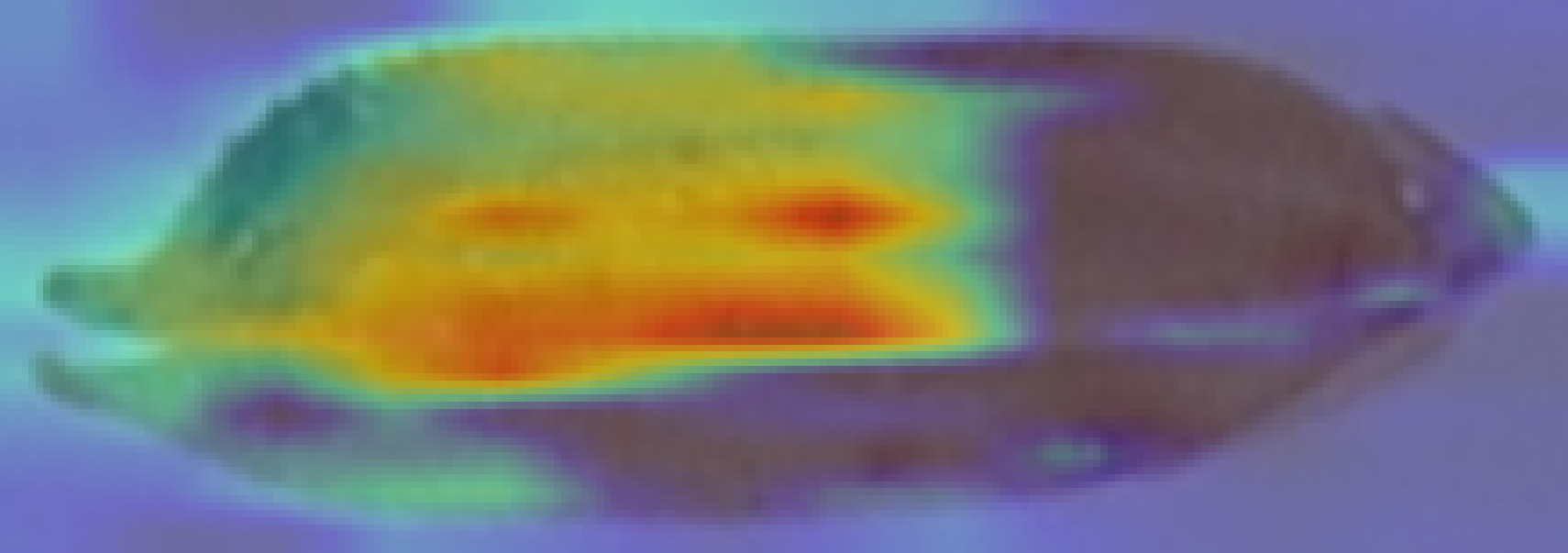}
        \caption{Vision transformer}
        \label{fig:vit_grad_cam}
    \end{subfigure}\hfill
    \begin{subfigure}[b]{0.245\textwidth}
        \centering
        \includegraphics[width=\linewidth]{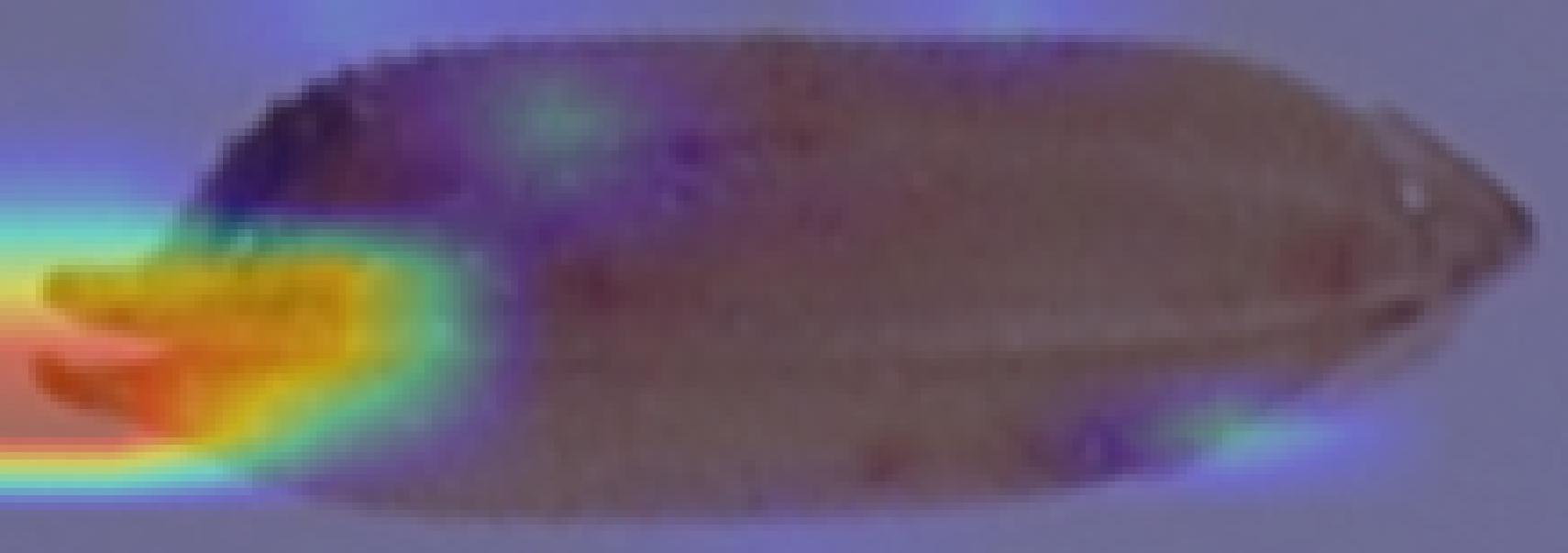}
        \caption{ConvNeXt}
        \label{fig:convnext_grad_cam}
    \end{subfigure}\hfill
    \begin{subfigure}[b]{0.245\textwidth}
        \centering
        \includegraphics[width=\linewidth]{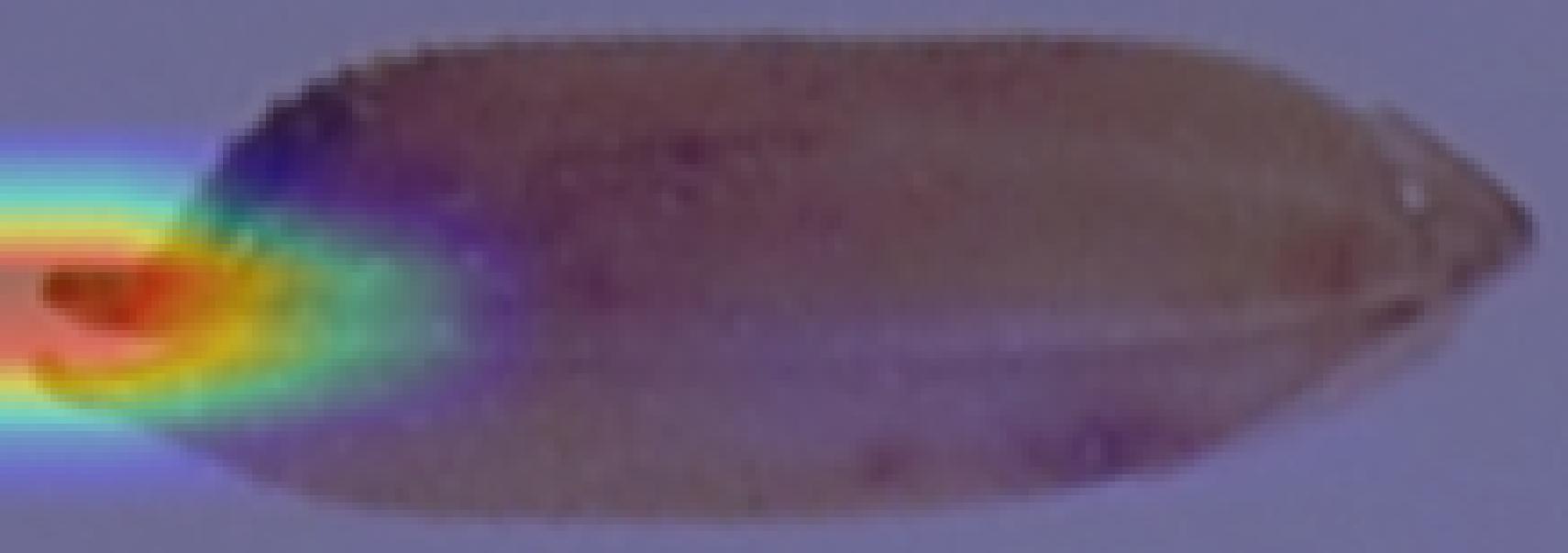}
        \caption{VGG16}
        \label{fig:vgg_grad_cam}
    \end{subfigure}
    
    \caption{Comparative visualization of attention maps on the \textbf{BC-15} rice variety, displaying the discriminative regions identified by the proposed distillation framework and benchmark models using Grad-CAM.}
    \label{fig:visualization_grad_cam}
\end{figure}

Beyond quantitative evaluation in terms of accuracy and computational efficiency, visual interpretability provides important insights into how knowledge distillation influences the attention behavior of the model. Specifically, it reveals which image regions and feature patterns are emphasized by the student model after learning from the teacher. To this end, Grad-CAM (Gradient-weighted Class Activation Mapping) was employed to generate class-discriminative heatmaps for the student, teacher, and pretrained baseline models. These visualizations offer an intuitive comparison of how different models attend to discriminative regions, particularly those associated with fine-grained textures and structural characteristics of rice grains. Figure~\ref{fig:visualization_grad_cam} illustrates representative attention maps for the BC-15 rice variety, highlighting differences in feature localization and attention consistency across models.

The effectiveness of the proposed hybrid distillation strategy is illustrated in the first row of Figure~\ref{fig:visualization_grad_cam}. The student main branch (Figure~\ref{fig:student_main_grad_cam}) exhibits attention patterns that closely align with those of the teacher main model (Figure~\ref{fig:teacher_main_grad_cam}), particularly around the grain tip, which represents a key discriminative region for rice variety identification. This alignment is evidenced by overlapping high-activation regions and can be attributed to the feature-based distillation process, which enforces representation similarity after the Global Average Pooling layer and guides the student to focus on the same informative feature channels as the teacher. 
Beyond this localized region, the student model demonstrates broader and more coherent attention coverage across the entire grain body, capturing fine-grained textures and structural details such as ridges and sharp boundaries. This behavior reflects the combined effect of feature-based distillation and the dense connectivity of the student architecture, which facilitates the propagation of low-level visual cues to deeper layers. A similar attention consistency is observed in the student auxiliary branch (Figure~\ref{fig:student_aux_grad_cam}), indicating that self-distillation contributes to stabilizing feature representations across network depths.

A comparison with pretrained convolutional baseline models further highlights the advantages of the proposed approach. MobileNetV2 (Figure~\ref{fig:mobileNet_grad_cam}), EfficientNetB0 (Figure~\ref{fig:efficientNet_grad_cam}), and ResNet18 (Figure~\ref{fig:resnet_grad_cam}) exhibit strongly localized attention concentrated primarily on the grain tip. While this region is critical for discrimination, their attention maps remain relatively narrow and show limited sensitivity to texture variations along the grain body when compared with the student model (Figure~\ref{fig:student_main_grad_cam} and Figure~\ref{fig:student_aux_grad_cam}). DenseNet121 (Figure~\ref{fig:densenet_grad_cam}) demonstrates improved preservation of low-level features due to dense connectivity, allowing it to attend to both the grain tip and boundary regions; however, its activation remains less comprehensive and consistent than that of the proposed student model.

In contrast, transformer-based architectures such as Swin-Tiny (Figure~\ref{fig:swin_grad_cam}) and ViT (Figure~\ref{fig:vit_grad_cam}) exhibit more scattered and fragmented attention patterns, failing to establish a clear focal region on the grain. The global self-attention mechanism distributes focus across multiple regions, which limits the ability of these models to capture fine-grained morphological details from the relatively small rice dataset and contributes to their inferior performance compared with the proposed CNN-based distillation framework.

These visualizations indicate that the proposed student model achieves a favorable balance between precise localization of key discriminative regions and holistic understanding of grain structure, resulting in more informative and consistent attention behavior than both convolutional and transformer-based benchmark models.

\subsection{Generalization performance on plant leaf disease datasets}

To further demonstrate the robustness and practical applicability of the proposed method, we extended our evaluation to four plant leaf disease datasets, including rice, potato, coffee, and corn. As summarized in Table~\ref{tab:generalization_results}, the experimental results indicate that the student model not only maintains performance closely comparable to the teacher but also surpasses the results reported in previous studies.

The most significant improvement is observed on the potato leaf disease dataset. Earlier studies reported considerably lower accuracies, with~\cite{Shabrina2024} achieving 73.63\% and~\cite{sujatha2025} reporting 62.60\%, while the more recent handcrafted feature fusion approach proposed by~\cite{HoangPhan2026} achieved 81.03\%. In contrast, the proposed student model attains an accuracy of 84.57\% on the main branch, improving upon the previously reported best result by more than 3.5\%. This performance gain reflects the strong capability of the proposed framework to discriminate subtle and complex disease patterns.

\begin{table}[H]
    \centering
    \footnotesize
    \caption{Generalization performance of the proposed method compared to previous studies across four plant leaf disease datasets.}
    \label{tab:generalization_results}

    \renewcommand{\arraystretch}{1.3}
    
    \newcolumntype{Y}{>{\centering\arraybackslash}X}

    \begin{tabularx}{\linewidth}{l Y Y Y Y}
    \toprule
    \multirow{2}{*}{\textbf{Study} / \textbf{Model}} & \multicolumn{4}{c}{\textbf{Accuracy (\%)}} \\ 
    \cmidrule(lr){2-5} 
     & \textbf{Rice leaf} & \textbf{Potato leaf} & \textbf{Coffee leaf} & \textbf{Corn leaf} \\ 
    \midrule
    \multicolumn{5}{l}{\textit{\textbf{Previous studies}}} \\
    \cite{HoangPhan2026} &  --  & 81.03 & --- & -- \\
    \cite{sujatha2025}   &  --  & 62.60 & --- & -- \\
    \cite{Meesala2025}   & 99.0 &   --  & --- & -- \\
    \cite{Yang2024}      &  --  &   --  & --- & 95.18 \\
    \cite{Shabrina2024}  &  --  & 73.63 & --- & -- \\
    \midrule
    \multicolumn{5}{l}{\textit{\textbf{Teacher model}}} \\
    Auxiliary branch & 99.58 & 85.53 & 99.53 & 95.23 \\
    Main branch      & 99.58 & 85.53 & 99.38 & 95.94 \\
    
    \midrule
    \multicolumn{5}{l}{\textit{\textbf{Student model}}} \\
    Auxiliary branch & \textbf{99.24} & \textbf{83.28} & \textbf{98.60} & \textbf{95.70} \\
    Main branch      & \textbf{99.07} & \textbf{84.57} & \textbf{98.91} & \textbf{95.70} \\
    \bottomrule
    \end{tabularx}
\end{table}

A similar trend is observed on the rice leaf disease dataset, where the student model achieves accuracies of 99.24\% on the auxiliary branch and 99.07\% on the main branch, both exceeding the 99.0\% accuracy reported in~\cite{Meesala2025}. On the corn leaf disease dataset, the student model reaches an accuracy of 95.70\% on both branches, surpassing the 95.18\% achieved by the improved YOLOv8-based approach proposed in~\cite{Yang2024}. Collectively, these results demonstrate the robustness and strong generalization capability of the proposed approach across diverse crop datasets.

For the coffee leaf disease dataset, direct comparisons with prior work are not available. Nevertheless, the achieved accuracy of 98.91\% provides a strong reference point for future studies. The performance gap between the student and teacher models on this dataset remains below 1.0\%, indicating that the proposed distillation strategy is highly stable and enables the compact student model to preserve nearly the full classification capability of the teacher, even when applied to plant species not encountered during the initial development.

The results summarized in Table~\ref{tab:generalization_results} demonstrate that the proposed framework achieves reliable performance across a wide range of plant leaf disease identification tasks. The strong generalization observed across all evaluated datasets suggests that the distillation strategy effectively transfers robust and transferable representations to the student model, highlighting its suitability for deployment in resource-constrained smart agriculture environments.

\section{Discussion}\label{sec:discussion}

This section discusses the underlying reasons behind the effectiveness of the proposed hybrid knowledge distillation framework, with a particular focus on architectural design, training dynamics, stability on small datasets, and inherent limitations of the teacher-student paradigm.

\paragraph{Insights into the proposed framework}
The proposed hybrid knowledge distillation framework demonstrates strong capability in constructing lightweight models that achieve high performance across diverse agricultural tasks. This effectiveness can be attributed to the careful design of a customized student architecture based on inverted residuals and dense connectivity, together with a multi-objective loss formulation. By jointly integrating hard-label supervision, feature-based alignment, response-based distillation, and self-distillation, the framework enables multi-level knowledge transfer, allowing the student model to capture both high-level semantic information and fine-grained textural details. As a result, the student learns a structured and discriminative feature space rather than merely replicating the teacher's final predictions.

\paragraph{Training dynamics and convergence behavior}
Analysis of the training dynamics reveals several advantages of the proposed method. Across multiple datasets, the student model closely follows the teacher during training, maintaining a minimal performance gap. This behavior is evident in Figure~\ref{fig:curves_rice_leaf} for the rice leaf disease dataset and Figure~\ref{fig:curves_coffee_leaf} for the coffee leaf disease dataset. Benefiting from the teacher's guidance, the student achieves a rapid reduction in training loss within the first ten epochs, as shown in Figures~\ref{fig:curves_rice_leaf}, \ref{fig:curves_coffee_leaf}, and \ref{fig:curves_corn_leaf}. Moreover, the student exhibits improved resistance to overfitting. On the potato leaf disease dataset (Figure~\ref{fig:curves_potato_leaf}), the gap between training and validation loss is substantially smaller for the student than for the teacher, indicating more stable generalization.

\paragraph{Limitations and challenges}
Despite these strengths, the framework faces certain limitations common to deep learning in data-constrained agricultural scenarios. First, training stability is reduced on small-scale datasets (rice variety, potato, and corn), where validation accuracy and loss show noticeable fluctuations (Figures~\ref{fig:curves_bc15_rice_variety}, \ref{fig:curves_potato_leaf}, \ref{fig:curves_corn_leaf}). These oscillations primarily arise from limited samples causing high mini-batch variance and less representative gradient updates, issues less pronounced on larger datasets like rice leaf disease. Second, as inherent to teacher-student paradigms, the student may inherit biases or errors from the teacher, observable in correlated performance variations on the corn and potato datasets (Figures~\ref{fig:curves_potato_leaf} and~\ref{fig:curves_corn_leaf}). The diverse supervisory signals (particularly hard-label supervision and self-distillation) help mitigate these effects, but complete elimination remains challenging.

Overall, the proposed approach achieves a robust balance between performance, efficiency, and practicality, making it highly suitable for resource-constrained smart agriculture applications. The compact student model delivers state-of-the-art results across varied tasks while enabling on-device deployment, a critical step toward scalable, real-time crop monitoring and disease diagnosis systems.

\section{Conclusion}\label{sec:conclusion}

This study proposed a novel hybrid knowledge distillation framework aimed at deploying lightweight yet high-performance models for smart agriculture applications. By designing a customized student architecture that combines inverted residual blocks with dense connectivity and adopting a hybrid multi-objective distillation strategy, including hard-label supervision, feature-based alignment, response-based distillation, and self-distillation, the proposed framework effectively transfers robust discriminative knowledge from a high-capacity teacher to a compact student model.

Extensive experiments on rice seed variety identification and multiple plant leaf disease datasets demonstrate that the distilled student model achieves near-teacher recognition performance while substantially reducing computational complexity and model size. The proposed student consistently matches or outperforms representative lightweight and heavyweight benchmark architectures in terms of accuracy–efficiency trade-off, while exhibiting strong generalization across diverse agricultural tasks. These results highlight the robustness and practical applicability of the proposed framework, particularly for deployment on resource-constrained edge devices in real-world smart agriculture scenarios.

Despite its effectiveness, the proposed framework relies on multiple loss components with associated weighting coefficients ($\lambda$) and a distillation temperature ($\tau$), which require careful tuning for optimal performance. Future work will explore automated hyperparameter optimization strategies to adaptively tune the loss weights and distillation temperature, further improving the robustness, stability, and deployment readiness of the student model across diverse agricultural scenarios.

\section*{Supplementary figures (learning curves):}\label{sec:appendix_figures}

\begin{figure}[H]
    \centering

    \begin{subfigure}{\textwidth}
        \centering
        \includegraphics[width=\linewidth]{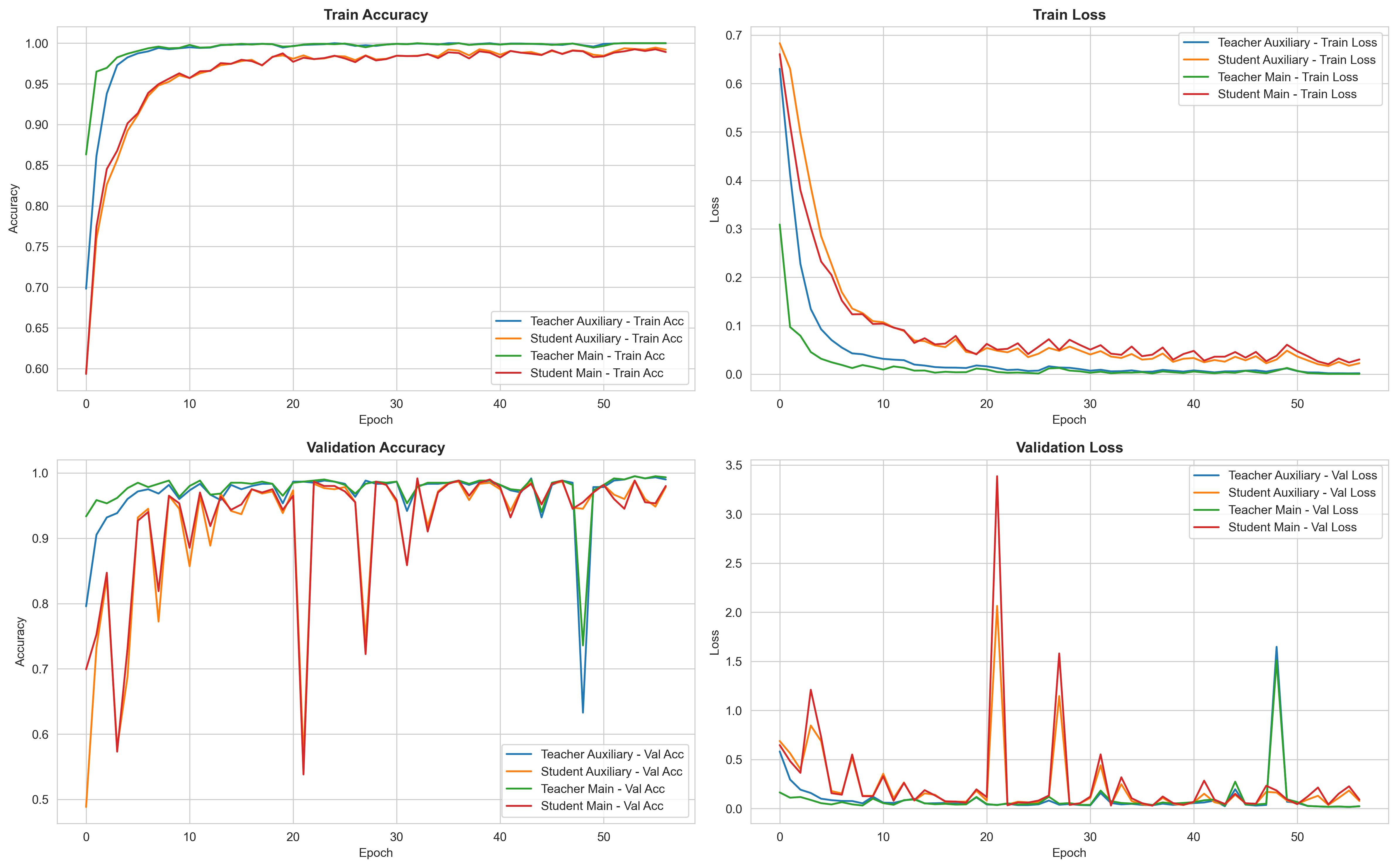}
        \caption{Training and validation learning curves on the BC-15 rice variety dataset.}
        \label{fig:curves_bc15_rice_variety}
    \end{subfigure}
    
    \caption{Training and validation learning curves for the teacher and student models across all five evaluated datasets (part 1).}
    \label{fig:all_learning_curves_part1}
\end{figure}

\begin{figure}[H]\ContinuedFloat
    \centering

    \begin{subfigure}{\textwidth}
        \centering
        \includegraphics[width=\linewidth]{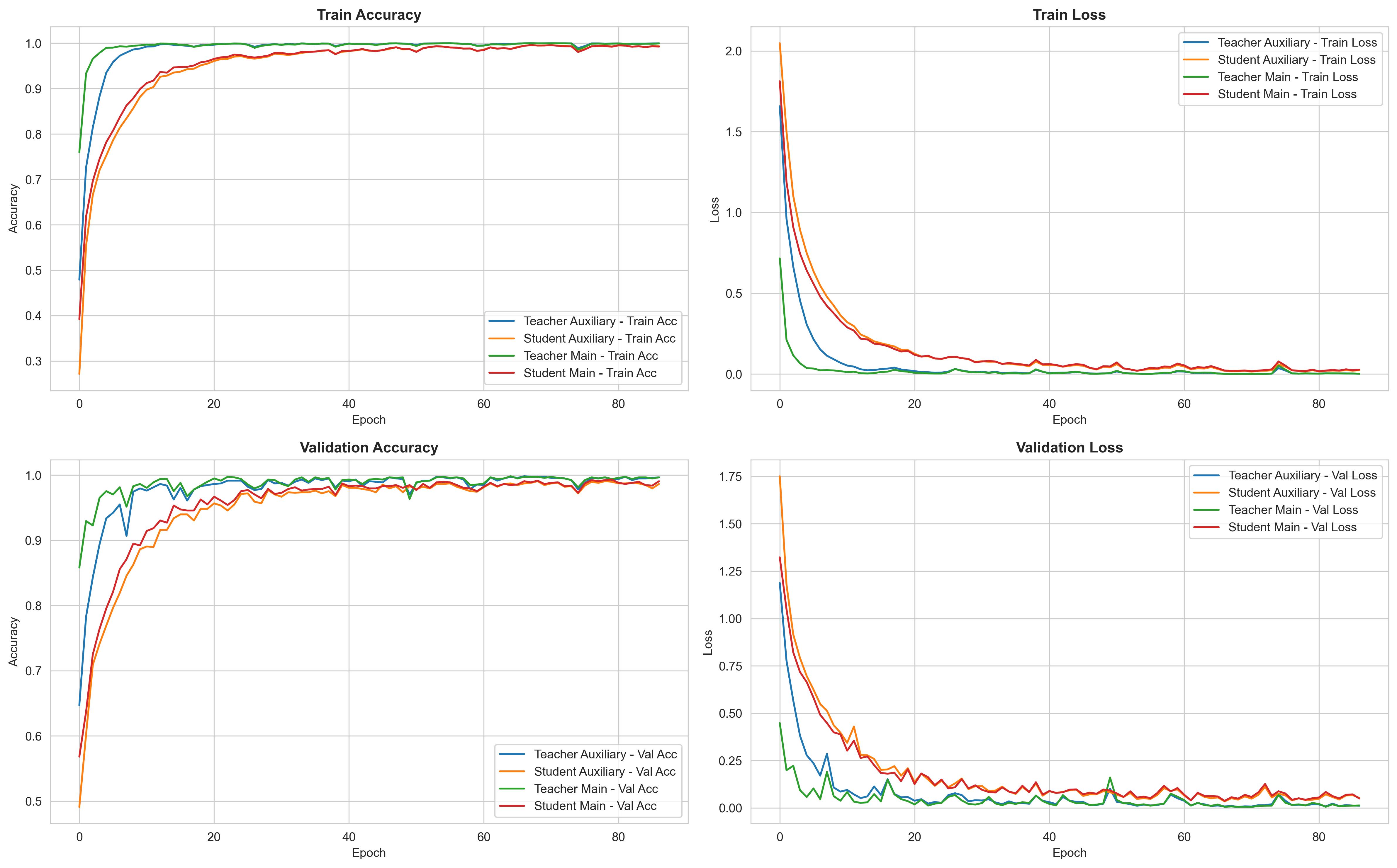}
        \caption{Training and validation learning curves on the rice leaf disease dataset.}
        \label{fig:curves_rice_leaf}
    \end{subfigure}

    \vspace{0.3cm}
    
    \begin{subfigure}{\textwidth}
        \centering
        \includegraphics[width=\linewidth]{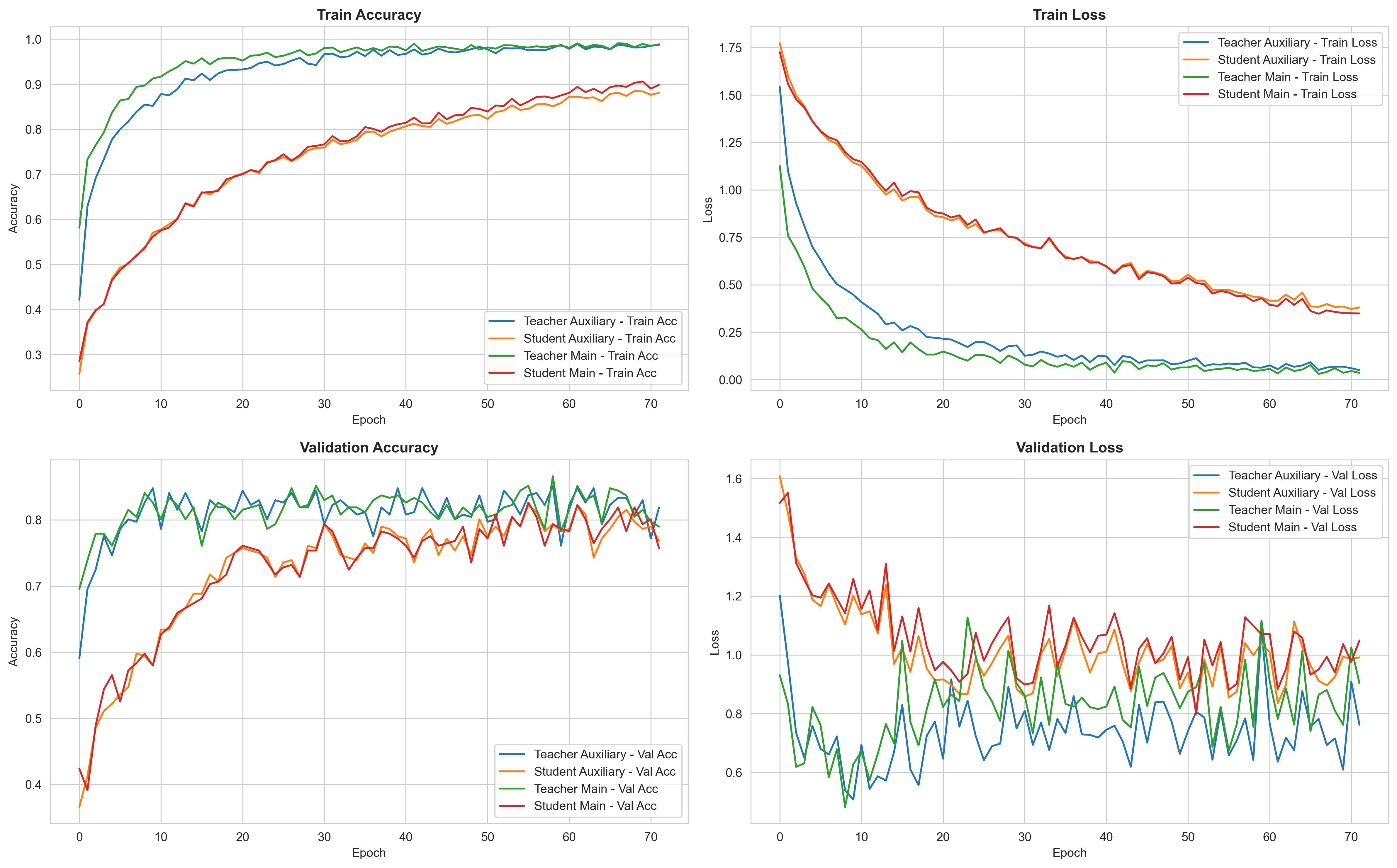}
        \caption{Training and validation learning curves on the potato leaf disease dataset.}
        \label{fig:curves_potato_leaf}
    \end{subfigure}
    
    \caption{Training and validation learning curves for the teacher and student models across all five evaluated datasets (part 2).}
    \label{fig:all_learning_curves_part2}
\end{figure}

\begin{figure}[H]\ContinuedFloat
    \centering

    \begin{subfigure}{\textwidth}
        \centering
        \includegraphics[width=\linewidth]{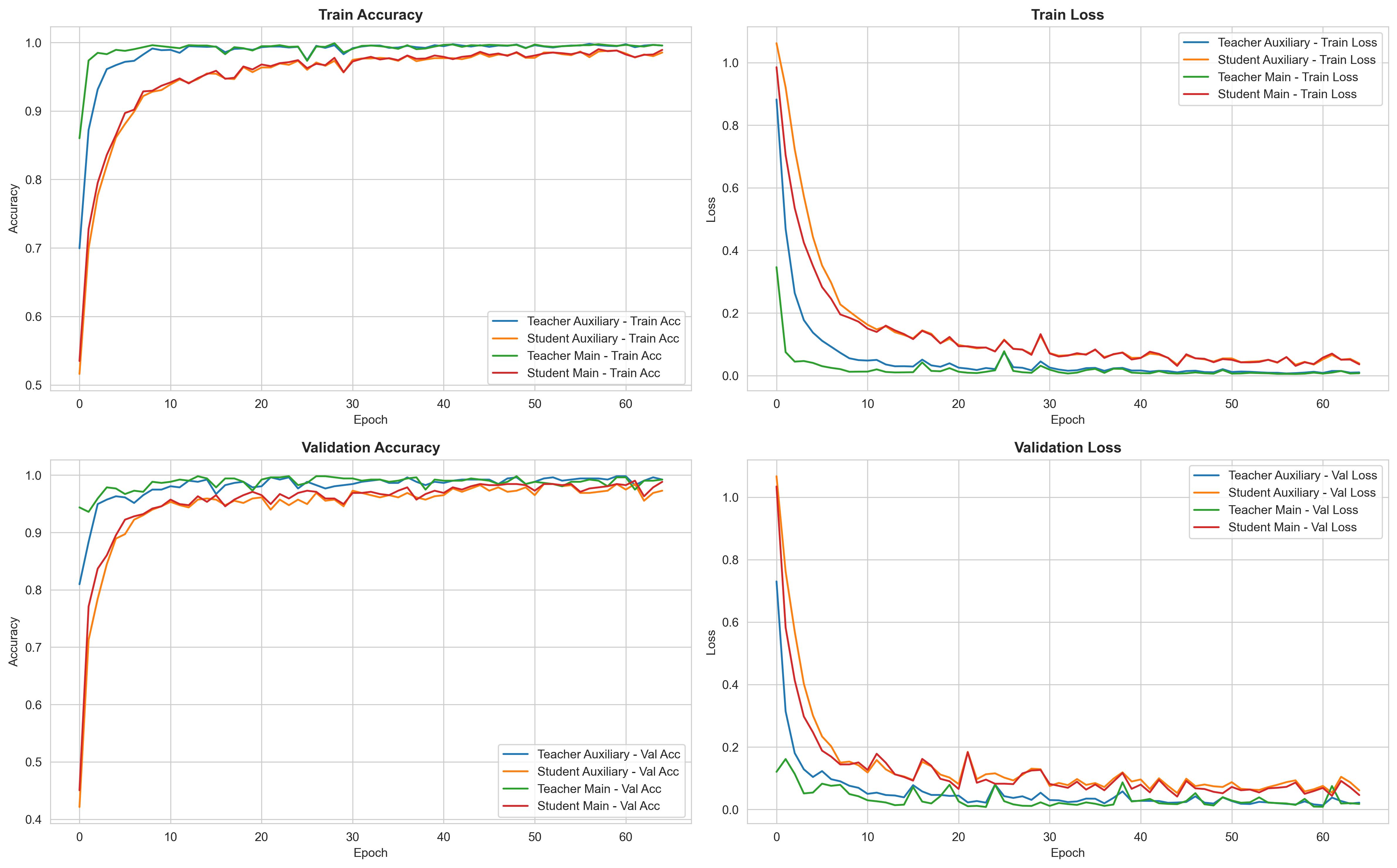}
        \caption{Training and validation learning curves on the coffee leaf disease dataset.}
        \label{fig:curves_coffee_leaf}
    \end{subfigure}

    \vspace{0.3cm}
    
    \begin{subfigure}{\textwidth}
        \centering
        \includegraphics[width=\linewidth]{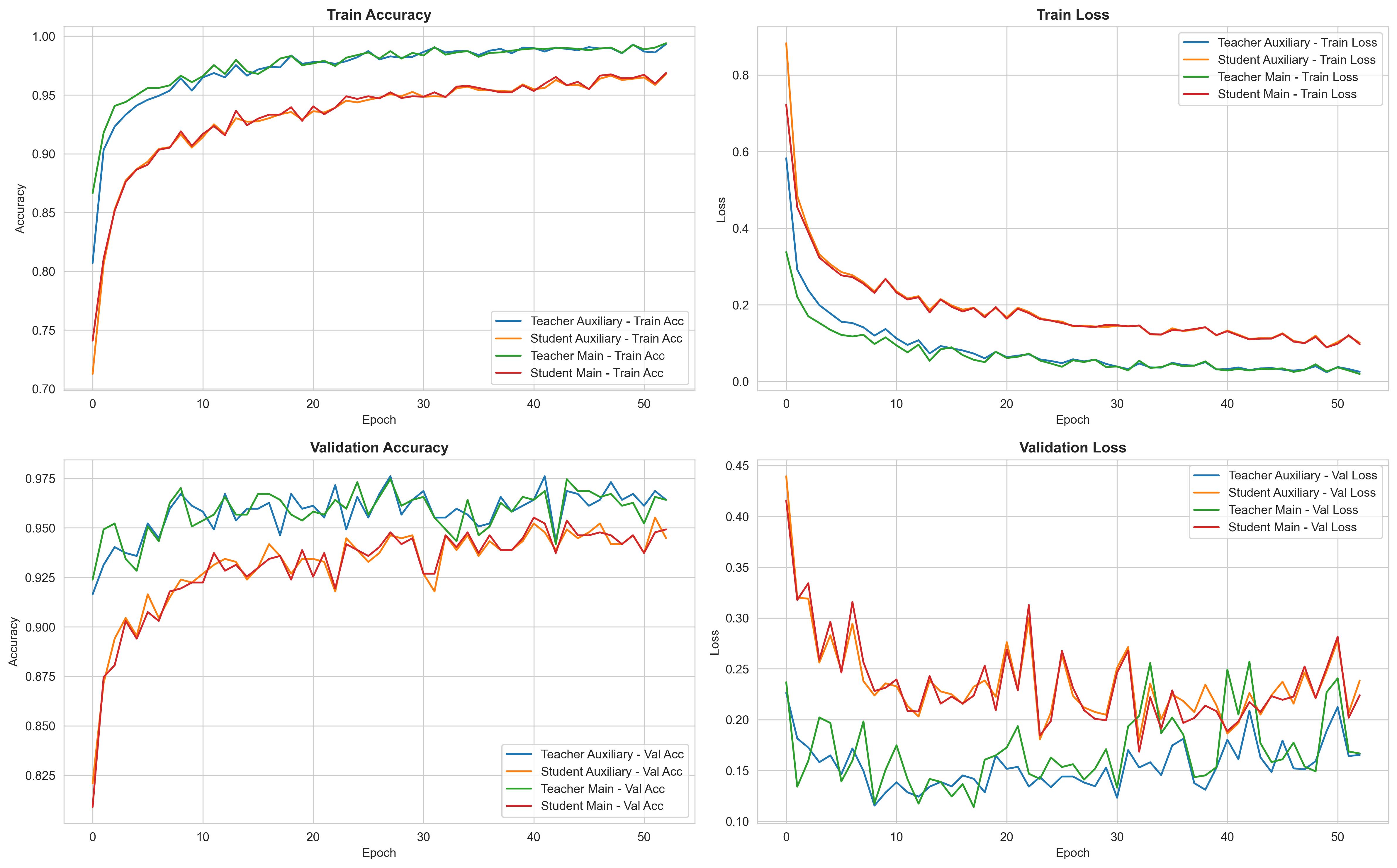}
        \caption{Training and validation learning curves on the corn leaf disease dataset.}
        \label{fig:curves_corn_leaf}
    \end{subfigure}
    \caption{Training and validation learning curves for the teacher and student models across all five evaluated datasets (part 3).}
    \label{fig:all_learning_curves_part3}
    
\end{figure}

\section*{Funding:} This research received no funding.




\section*{Declaration of interests:}
 
The authors declare that they have no known competing financial interests or personal relationships that could have appeared to influence the work reported in this paper.

\section*{Declaration of generative AI and AI-Assisted technologies in the writing process:}

During the preparation of this manuscript, Grammarly and generative large language models were employed to improve grammar and refine wording for clarity. The authors subsequently reviewed and revised the content as necessary and take full responsibility for the final version of the manuscript.

\bibliographystyle{elsarticle-harv}
\bibliography{Main/KD_refers}

\end{document}